\documentclass[lettersize,journal]{IEEEtran}
\usepackage{amsmath,amssymb}
\usepackage{psfrag}
\usepackage{epsfig}
\usepackage{cite}
\usepackage{graphics}
\usepackage{color}
\usepackage{cases}
\usepackage{subfigure}
\usepackage{epstopdf}
\usepackage{epsfig}
\usepackage{hyperref}
\usepackage{multirow}
\usepackage{mathrsfs}
\usepackage{algorithmic}
\usepackage{algorithm}
\usepackage{url}

\usepackage{booktabs}
\usepackage{pifont}

\usepackage{paracol}
\usepackage[table]{xcolor}
\usepackage{colortbl}
\renewcommand{\vec}[1]{\boldsymbol{#1}}
\newcommand{\cmark}{\ding{51}}%
\newcommand{\xmark}{\ding{55}}%

\begin{document}

\title{Improving Fast Adversarial Training Paradigm: An Example Taxonomy Perspective}

\author{Jie Gui,~\IEEEmembership{Senior Member,~IEEE,} Chengze Jiang,~\IEEEmembership{Student Member,~IEEE,} Minjing Dong, Kun Tong, Xinli Shi,~\IEEEmembership{Senior Member,~IEEE,} Yuan Yan Tang,~\IEEEmembership{Life Fellow,~IEEE,} Dacheng Tao,~\IEEEmembership{Fellow,~IEEE}
\thanks{J. Gui is with the School of Cyber Science and Engineering, Southeast University and with Purple Mountain Laboratories, Nanjing 210000, China (e-mail: guijie@seu.edu.cn).}
\thanks{C. Jiang, K. Tong, and X. Shi are with the School of Cyber Science and Engineering, Southeast University, Nanjing 210000, China (e-mail: czjiang@seu.edu.cn; xytk2000@163.com; xinli$\_$shi@seu.edu.cn).}
\thanks{M. Dong is with the Department of Computer Science, City University of Hong Kong. (e-mail: minjdong@cityu.edu.hk).}
\thanks{Y. Tang is with the Department of Computer and Information Science, University of Macau, Macau 999078, China (e-mail: yytang@um.edu.mo).}
\thanks{D. Tao is with the College of Computing $\&$ Data Science at Nanyang Technological University, $\#$32 Block N4 $\#$02a-014, 50 Nanyang Avenue, Singapore 639798 (email: dacheng.tao@gmail.com).}
}

\markboth{Journal of \LaTeX\ Class Files,~Vol.~14, No.~8, August~2021}%
{Shell \MakeLowercase{\textit{et al.}}: A Sample Article Using IEEEtran.cls for IEEE Journals}


\maketitle

\begin{abstract}
While adversarial training is an effective defense method against adversarial attacks, it notably increases the training cost. To this end, fast adversarial training (FAT) is presented for efficient training and has become a hot research topic. However, FAT suffers from catastrophic overfitting, which leads to a performance drop compared with multi-step adversarial training. However, the cause of catastrophic overfitting remains unclear and lacks exploration. 
In this paper, we present an example taxonomy in FAT, which identifies that catastrophic overfitting is caused by the imbalance between the inner and outer optimization in FAT. Furthermore, we investigated the impact of varying degrees of training loss, revealing a correlation between training loss and catastrophic overfitting. Based on these observations, we redesign the loss function in FAT with the proposed dynamic label relaxation to concentrate the loss range and reduce the impact of misclassified examples. Meanwhile, we introduce batch momentum initialization to enhance the diversity to prevent catastrophic overfitting in an efficient manner. Furthermore, we also propose Catastrophic Overfitting aware Loss Adaptation (COLA), which employs a separate training strategy for examples based on their loss degree. Our proposed method, named example taxonomy aware FAT (ETA), establishes an improved paradigm for FAT. Experiment results demonstrate our ETA achieves state-of-the-art performance. Comprehensive experiments on four standard datasets demonstrate the competitiveness of our proposed method.
\end{abstract}

\begin{IEEEkeywords}
Fast Adversarial Training, Robustness, Adversarial Example, Catastrophic Overfitting.
\end{IEEEkeywords}

\section{Introduction}
\IEEEPARstart{D}{eep} learning has achieved remarkable performance on computer vision \cite{SurViT, Vivit, Swin, DCMPZH}. Nonetheless, it has been reported that deep learning methods are vulnerable to adversarial attacks \cite{CBUAA, yin2023push, BIM}, which can lead to incorrect predictions of models or even manipulate the predictions of models \cite{QFAASD, AAMEAA, UNIAP}. To mitigate the risks, defense approaches are proposed and investigated \cite{AADefense, PAAD, defenseSur}. Among them, adversarial training is proven the most effective method in strengthening neural networks against attacks \cite{MART, zhuo2022self}. Adversarial training can be categorized into two groups depending on the approach employed to produce adversarial examples (AEs) for training i.e., multi-step and fast adversarial training (FAT) \cite{TDAT, PGIMEP}. Multi-step adversarial training significantly enhances the adversarial robustness of deep learning models \cite{wu2022towards}. However, it requires additional time for computing gradients to generate AEs for training \cite{GradAlign}. To alleviate computational burdens, FAT adopts the fast gradient sign method (FGSM) to acquire AEs for training, garnering considerable interest for their effectiveness and efficiency \cite{FGSMRS, TDAT, GAT}.

\begin{figure}[t]
	\centering
	\includegraphics[scale=0.345]{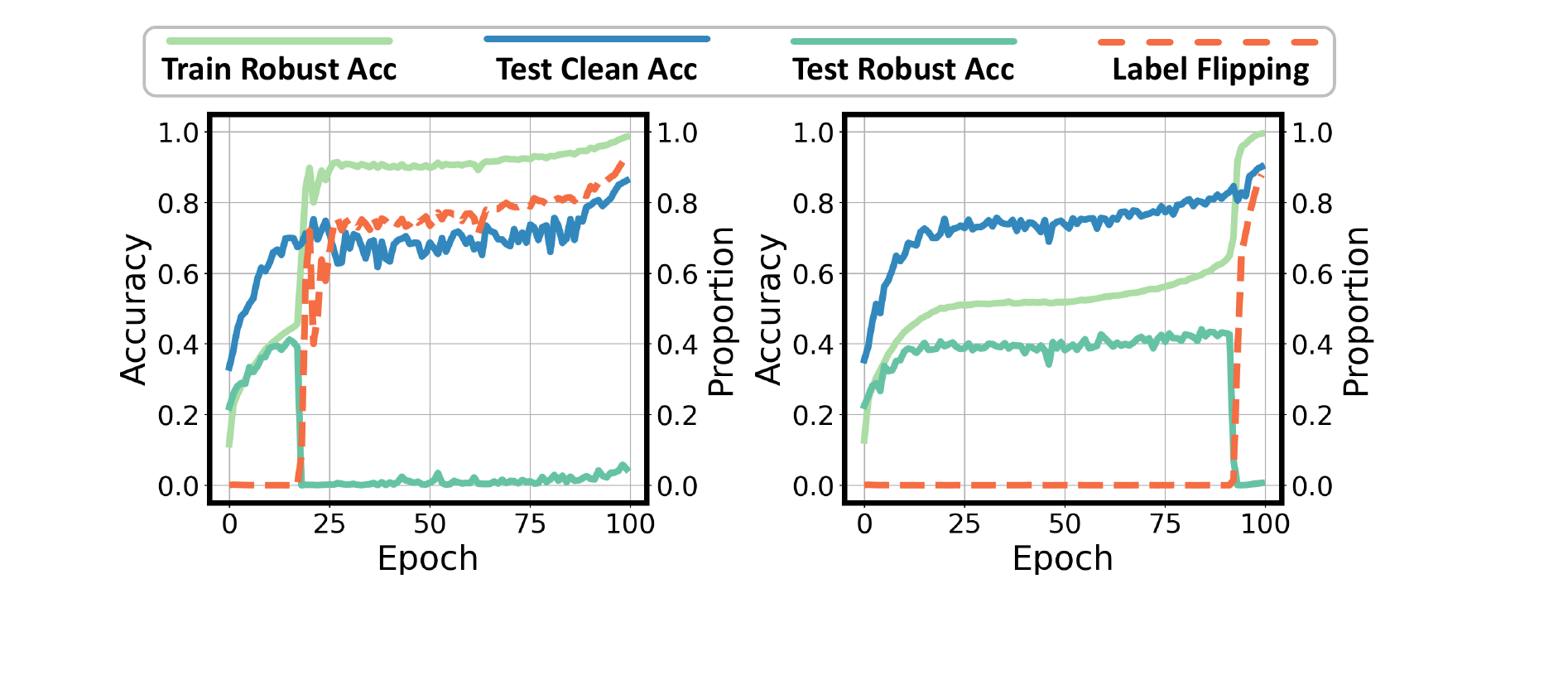}
	\caption{Catastrophic overfitting and label flipping. We adopt the ResNet18 to perform FGSM-AT \cite{FGSM} and FGSM-RS \cite{FGSMRS} on the CIFAR-10 dataset with perturbation budget $\epsilon=8/255$. The train and test robust accuracy is evaluated by the FGSM with $\alpha=8/255$ and PGD-10 with $\alpha=2/255$. The red line denotes the proportion of Case 4 (as Fig. \ref{NumCO}) to all misclassified examples, which infers the proportion of label flipping examples explodes once catastrophic overfitting occurs.}
	\label{COCase}
\end{figure}

However, FAT encounters a significant problem known as catastrophic overfitting \cite{stableFat}. This problem causes the adversarial robustness of the model to be compromised on the test set against multi-step adversarial attacks \cite{FATSC}, as illustrated in Fig. \ref{COCase}. Research demonstrates that the issue of catastrophic overfitting occurring during training is related to a sharp decline in the generalization of neural networks \cite{stableFat, SATDS}. To this end, studies have been devoted to investigating and addressing this problem during past years \cite{OFBAT}. One approach to handle catastrophic overfitting is FGSM with random initialization (FGSM-RS) combined with an early stopping \cite{FGSMRS}. Moreover, Andriushchenko $et~al.$ demonstrate the connection between neural network structure and catastrophic overfitting, thereby introducing the regularization for developing gradient align (GradAlign) \cite{GradAlign}. After that, the cause of catastrophic overfitting is explained from the perspective of the decision boundary and loss surface, leading to the stable FAT for preventing catastrophic overfitting \cite{stableFat}. Besides, the knowledge distillation is leveraged for logit smoothing to mitigate this problem \cite{split}. Recently, there are some new perspectives and solutions are presented to prevent catastrophic overfitting, containing prior-guided initialization (FGSM-MEP) \cite{PGIMEP}, guided adversarial training (GAT) \cite{subspace}, Noise FGSM (NFGSM) \cite{NFGSM}, etc. However, these variants do not adequately explain the reasons for catastrophic overfitting, particularly from the perspectives of adversarial examples and training loss changes. Furthermore, the impact patterns of misclassified examples on FAT and catastrophic overfitting are not investigated. Given these problems in FAT, it is valuable to perform systematic analysis on these issues and develop tailored improvements to enhance training performance and eliminate catastrophic overfitting. Thus, in this paper, we propose to explore the example taxonomy in FAT to reveal the causes of catastrophic overfitting from a perspective of optimization imbalance. Based on the analysis, we introduce Example Taxonomy Aware FAT (ETA) to tackle these issues efficiently. The main contributions of this paper are summarized as follows:
\begin{itemize}
    \item We present a novel taxonomy of training examples to understand catastrophic overfitting. The results reveal that the change of example numbers in different cases can reveal the cause of catastrophic overfitting.
    \item To analyze catastrophic overfitting during optimization, we investigate the variation in the number of examples with different loss magnitudes and reveal the necessity of the concentration of loss range.
    \item From the observation results, we present a new FAT paradigm, which improves perturbation initialization, label relaxation, loss function, and loss adjustment.
    \item Extensive experiments are performed on four standard datasets for comparison. The results demonstrate that our ETA outperforms FAT and multi-step adversarial training while eliminating catastrophic overfitting.
\end{itemize}
Section \ref{RW} presents a novel taxonomy of examples with mathematical definitions, analyzing the relationship between the number of examples, their losses, and the catastrophic overfitting through this taxonomy. Subsequently, implementation details of the proposed ETA are provided in Section \ref{ProblemDes}. Section \ref{EandA} presents the results of systematic experiments and offers relevant analyses. Section \ref{ConOut} concludes the paper and provides an outlook on future research directions. We hope our work will inspire a deeper understanding of FAT and enhance the adversarial robustness of deep learning models.
\par
This paper is the journal extension of our conference paper \cite{TDAT}. Significant improvements and extensions are made compared to our original conference version. The main differences are summarized in four aspects: 1) Compared to the conference version, we provide a more formal and precise definition of the example taxonomy, thereby offering a clearer understanding of catastrophic overfitting in subsection \ref{EPT}. We also analyze the catastrophic overfitting from the training loss perspective in subsection \ref{CBTL}. We investigate the relationship between training loss, catastrophic overfitting, and model accuracy. Our results present a deeper understanding and analysis of catastrophic overfitting. 2) Inspired by our findings in this paper, we propose ETA in section \ref{Met}, which optimizes example loss adaptation to mitigate the impact of misclassified examples on training. Our ETA achieves better robustness against multi-step adversarial attacks while preventing catastrophic overfitting. 3) More experiments involving comparisons of our ETA with state-of-the-art solutions, ablation studies, and performance analysis are provided in section \ref{EandA}. Specifically, we provide the performance comparison of ETA in subsection \ref{CEA}, efficiency analysis in subsection \ref{EffAnaSec}, ablation studies in subsection \ref{AblationSty}, hyperparameters effect in subsection \ref{EFCOLA}, and visualization in subsection \ref{VAResults}. In addition, we also provide experiments in subsection \ref{CwithEM} to improve the existing methods by using ETA as a plug-in. The results show that it can simultaneously improve the clean and robust accuracy of the existing methods. 4) We thoroughly rewrite the abstract, introduction, methodology, experiment, and conclusion to present a more comprehensive overview of motivation and approach. Furthermore, we refine all the figures and tables.

\section{Related Work}\label{RW}
\subsection{Adversarial Training}
Deep learning and neural networks have achieved significant progress in many fields \cite{GJSSSL, SOMA, DMVA, HHEAA}. However, they remain susceptible to adversarial attacks. Adversarial training adopts AEs as training data to enhance the adversarial robustness of deep neural networks to defend against adversarial attacks \cite{liu2022mutual}. Define dataset with label $(\vec{x}, \vec{y})$ follows distribution $\mathcal{D}$ contains $m$ classes and $n$ examples. The optimization objective of adversarial training involves internal maximization and external minimization \cite{ASTAT}:
\begin{equation}\label{ATDefine}
	\min_{\theta} \mathbb{E}_{(\vec{x}, \vec{y})\sim\mathcal{D}}\Big{[}\max_{\|\vec{\delta}\|_{\text{p}}\leq \epsilon} \mathcal{L}\big{(}f_{\theta}(\vec{x}+\vec{\delta}), \vec{y}\big{)}\Big{]},
\end{equation}
where $f_\theta(\cdot)$ denotes the model with parameter $\theta$, $\mathcal{L}$ represents the loss function, $\epsilon$ is the perturbation budget, and $\vec{\delta}$ signifies the adversarial perturbation with $\|\cdot\|_{\text{p}}$ denotes the $p$-norm \cite{kuang2024defense}. Internal maximization is achieved by generating worst-case AEs, while external minimization requires the model to classify the AEs correctly \cite{CFA, WAT}. Multi-step adversarial training adopts the projected gradient descent (PGD) to generate AEs \cite{PGD, BagAT}, which iteration formula is as follows:
\begin{equation}
	\vec{x}^{k+1}_{\text{adv}}=\Pi_{\epsilon}\Big{(}\vec{x}^{k}_{\text{adv}}+\alpha\cdot\text{sign}\big{(}\nabla_{\vec{x}^{k}_{\text{adv}}}\mathcal{L}(f_\theta(\vec{x}^{k}_{\text{adv}}),\vec{y})\big{)}\Big{)},
\end{equation}
where $t$ denotes the iteration step and $\alpha$ represents the step size. The projection function $\Pi_{\epsilon}(\cdot)$ guarantees that the adversarial perturbation remains bounded in the feasible region \cite{DAFA}. Many methods are presented to improve multi-step adversarial training, including regularization \cite{theoretically}, class balancing \cite{CFA, UATCW}, and attack strategies \cite{FATASS}. Consequently, multi-step adversarial training is shown to possess competitive ability in enhancing the robustness of neural networks \cite{LBGAT, theoretically}. However, the time overhead required for multi-step adversarial training is significantly higher than normal training due to PGD requiring multiple backpropagation computations \cite{NuAT}.

\begin{figure*}[t]\centering
	\includegraphics[scale=0.62]{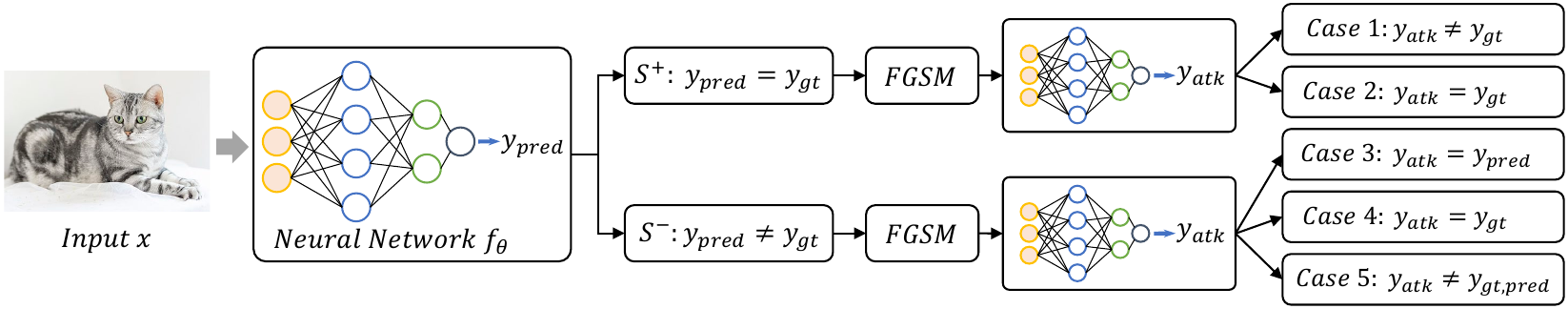}
	\caption{The proposed taxonomy of training examples in FAT. The five cases in this figure correspond to Case 1 to Case 5, from top to bottom.}
	\label{TabelSplit}
\end{figure*}

\subsection{Fast Adversarial Training and Catastrophic Overfitting}
To reduce the computational time of adversarial training, FAT develops FGSM instead of PGD to generate AEs, thus requiring only one additional backpropagation computation \cite{SDI}. The form of FGSM adopted in FAT is formulated as
\begin{equation}\label{FGSMFAT}
	\vec{x}_{\text{adv}} = \vec{x}+ \Phi_{\epsilon}\Big{(}\vec{\delta}_0 + \epsilon \cdot \text{sign}\big{(}\nabla_{\vec{x}}\mathcal{L}(f_{\theta}(\vec{x}+\vec{\delta}_0), \vec{y})\big{)}\Big{)},
\end{equation}
where $\vec{\delta}_0$ denotes the initialization perturbation. In the situation of $\Phi_{\epsilon}(\cdot)$ being the identity function and $\vec{\delta}_0=\vec{0}$, the formula is the definition of original FGSM \cite{NFGSM}. However, performing FAT with the original FGSM encounters catastrophic overfitting \cite{stableFat}. This problem results in training instability and even the breakdown of the training procedure. When this intriguing and confusing question occurs, the robust accuracy of the model on the training set increases dramatically, while the robust accuracy on the test set is almost destroyed \cite{OMDNRCO}. At the same time, the clean accuracy remains nearly no fluctuation. This problem has garnered widespread attention. FGSM-RS is presented to increase the diversity of AEs by sampling initial perturbations $\vec{\delta}_0$ from a uniform distribution $\mathcal{U}(-\epsilon, \epsilon)$. The results indicate that this approach can mitigate the catastrophic overfitting \cite{FGSMRS}. Subsequently, the sample-dependent adversarial initialization FGSM (FGSM-SDI) develops a learnable initialization scheme to enhance the initialization perturbations diversity \cite{SDI}. To investigate the impact of perturbation amplitude $\epsilon$ on FAT. The NFGSM addresses catastrophic overfitting by eliminating the projection operator $\Phi_{\epsilon}(\cdot)$ and amplifying the intensity of perturbation \cite{NFGSM}. Jia $et~al.$ present the FGSM-MEP to enhance the diversity of AEs and eliminate the catastrophic overfitting. This method uses historical information from previous training epochs to guide the initialization of perturbations \cite{PGIMEP}. Additionally, several methods aim to improve FAT by introducing regularization, thereby optimizing the trade-off between robust and clean accuracy \cite{GradAlign, GAT, NuAT}. Among these, GradAlign addresses the catastrophic overfitting by making the loss gradients with respect to AEs and clean examples closer during training, thereby achieving better robust accuracy \cite{GradAlign}. Meanwhile, GAT introduces a relaxation term to minimize the discrepancy between the model outputs for AEs and clean examples, aiming to achieve refined approximations of the loss \cite{GAT}. However, these methods do not adequately explain the mechanism behind catastrophic overfitting, and further improvements in robust accuracy are valuable. 

\subsection{Example Exploitation in Adversarial Training}
Example exploitation adopts the characters played by different examples to identify their influence and present enhancements. Specifically, to improve the robust generalization, self-adaptive training is presented, which dynamically adjusts the label of each example in training iteration and incorporates the model outputs into the method \cite{SATBER}. Misclassification Aware AdveRsarial Training (MART) investigates the impact of correctly classified and misclassified examples in multi step adversarial training, highlighting the importance of misclassified examples for robustness. Based on this, MART introduces a loss-gain term to exploit the two categories of examples in a differentiated manner \cite{MART}. Zhang $et~al.$ present a method that designs adaptive attack schemes for each example, selecting AEs with smaller losses to improve model robustness \cite{AWKTM}.

\section{Problem Description and Analysis}\label{ProblemDes}

\subsection{The Proposed Example Taxonomy}\label{EPT}
Catastrophic overfitting is identified as the overfitting of the model to AEs \cite{OMDNRCO}. Nonetheless, this explanation does not clarify why there is a significant drop in robust accuracy and a minimal effect on clean accuracy in the test dataset. Consequently, these issues drive us to develop a new perspective on investigating. Afterward, we present a taxonomy that categorizes the examples into five groups based on whether the examples are correctly classified before being attacked and whether these examples are successfully attacked. Then, we analyze the changes in the number of examples in each category. First, divide the training examples into two categories based on whether they are correctly classified:
\begin{equation}
	\begin{aligned}
		&S^+=\{f_\theta(\vec{x}) = \vec{y}\},\\
		&S^-=\{f_\theta(\vec{x}) \ne \vec{y}\},
	\end{aligned}
\end{equation}
where $S^+$ denotes the set of examples correctly classified by the model $f_\theta(\cdot)$, and $S^-$ represents the set of examples misclassified by the model $f_\theta(\cdot)$. Then, divide the $S^+$ set into two cases based on whether the AEs are misclassified:
\begin{subequations}
	\begin{align}
		&S^+_{\text{case 1}}=\{f_\theta(\vec{x}) = \vec{y} \cap f_\theta(\vec{x}_{\text{adv}})\ne \vec{y}\}, \label{CaseSplitOne} \\
		&S^+_{\text{case 2}}=\{f_\theta(\vec{x}) = \vec{y} \cap f_\theta(\vec{x}_{\text{adv}})= \vec{y}\}. \label{CaseSplitTwo}
	\end{align}
\end{subequations}
Then, based on whether the misclassified clean examples are successfully attacked and the case of the model classifies the corresponding AEs, divide the $S^-$ set into three cases as
\begin{subequations}
	\begin{align}
		&S^-_{\text{case 3}}=\{f_\theta(\vec{x}) \ne \vec{y} \cap f_\theta(\vec{x}_{\text{adv}})= f_\theta(\vec{x})\}, \label{CaseSplitThree} \\
		&S^-_{\text{case 4}}=\{f_\theta(\vec{x}) \ne \vec{y} \cap f_\theta(\vec{x}_{\text{adv}})= \vec{y}\}, \label{CaseSplitFour} \\
		&S^-_{\text{case 5}}=\{f_\theta(\vec{x}) \ne \vec{y} \cap f_\theta(\vec{x}_{\text{adv}})\ne \vec{y} \cap f_\theta(\vec{x}_{\text{adv}})\ne f_\theta(\vec{x})\}, \label{CaseSplitFive}
	\end{align}
\end{subequations}
Then, to provide a detailed explanation of the meanings of the five example cases, we provide the definitions in detail as
\begin{itemize}
	\item $S^+_{\text{case 1}}$: Model $f_\theta(\cdot)$ correctly classifies the clean example but misclassifies the corresponding AE.
	\item $S^+_{\text{case 2}}$: Model $f_\theta(\cdot)$ correctly classifies the clean example and also correctly classifies the corresponding AE.
	\item $S^-_{\text{case 3}}$: Model $f_\theta(\cdot)$ misclassifies the $i$-class clean example as $j$-class example, and classify the corresponding AE as $j$-class example.
	\item $S^-_{\text{case 4}}$: Model $f_\theta(\cdot)$ misclassifies the $i$-class clean example as $j$-class example, and classify the corresponding AE as $i$-class example.
	\item $S^-_{\text{case 5}}$: Model $f_\theta(\cdot)$ misclassifies the $i$-class clean example as $j$-class example, and classify the corresponding AE as $h$-class example with $h\ne i,j$.
\end{itemize}
To ensure presentation clarity and conciseness, we refer to the above five cases as cases 1 to 5.

\begin{figure}[t]\centering
	\includegraphics[scale=0.265]{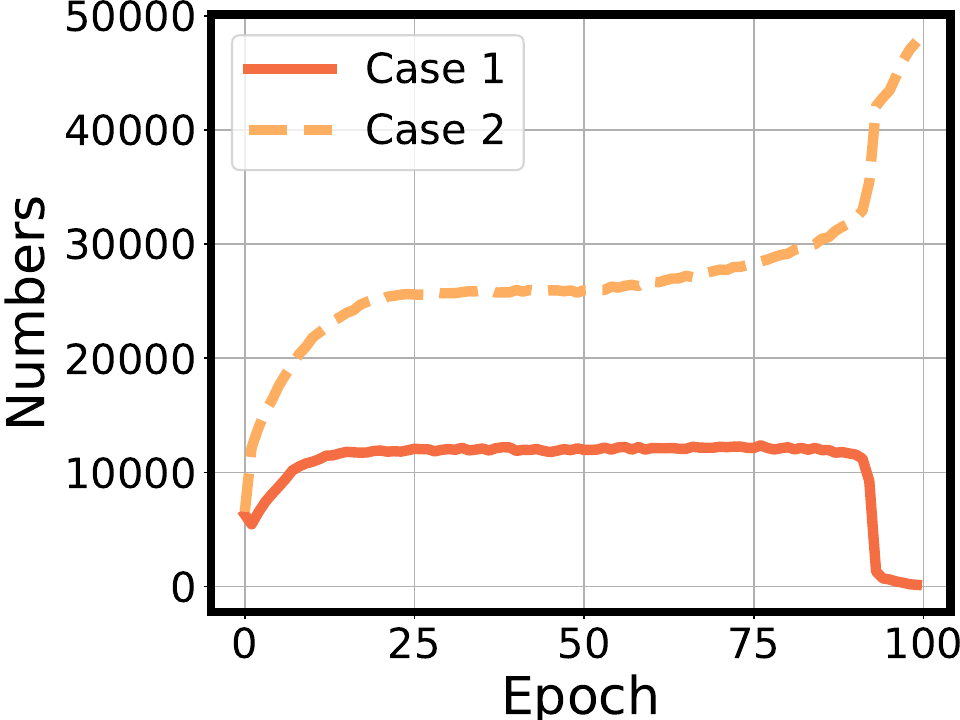}
	\includegraphics[scale=0.265]{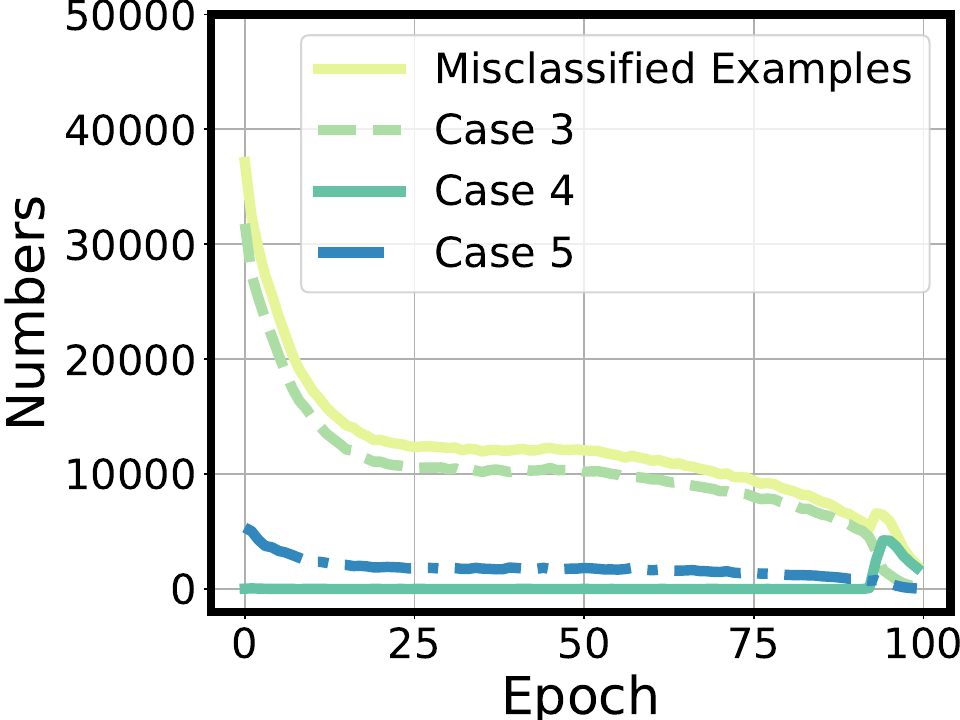}
	\caption{Numbers of the five cases during training. The sum of the five cases for each epoch equals the number of examples in the training dataset.}
	\label{NumCO}
\end{figure}

\subsection{Taxonomy and Catastrophic Overfitting}\label{TCO}
Figure \ref{TabelSplit} presents our example taxonomy. To illustrate the example taxonomy during optimization, we perform FGSM-RS \cite{FGSMRS} using ResNet-18 on the CIFAR10 with the perturbation budget $\epsilon=8/255$. Figure \ref{NumCO} illustrates the variation in the number of different cases over training. Specifically, as demonstrated in Fig. \ref{NumCO}(Left), in the $91$ epoch, the AEs generated by FGSM demonstrate a low attack success rate when catastrophic overfitting is present (as shown by case 1). That is to say, AEs lost their adversarial features and became insufficient to support the execution of FAT. Meanwhile, as observed in Fig. \ref{NumCO}(Right) and Fig.\ref{COCase}(Right), we observe an interesting phenomenon of \textbf{label flipping}, wherein a majority of misclassified clean examples are attacked to ground truth. In this situation, the internal maximization optimization in FAT as formula \eqref{ATDefine} becomes ineffective. AEs are obtained to satisfy the goal of the outer minimization problem rather than their original objective, resulting in the abrupt failure of FAT. Conversely, a significant portion of failed attack examples (case 2) and nearly harmless AEs (case 4) mislead the model, causing it to focus on the distribution around clean examples. As a result, the clean accuracy on the test dataset is minimally changed, but the robustness of the model is collapsed. In general, within the optimization framework \eqref{ATDefine}, we observe an imbalance between the inner and outer optimization, where the minimization surpasses the maximization. Therefore, it is crucial to augment maximization by leveraging the diversity of AEs while concurrently refining the minimization problem to extract knowledge from these AEs better.

\begin{figure}[t]\centering
	\includegraphics[scale=0.265]{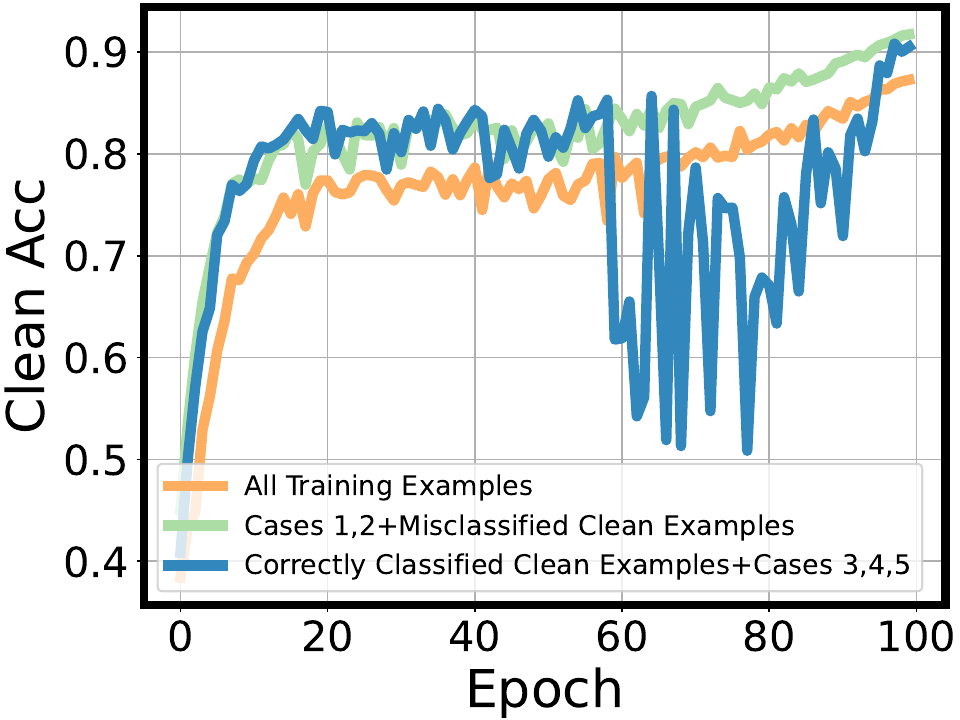}
	\includegraphics[scale=0.265]{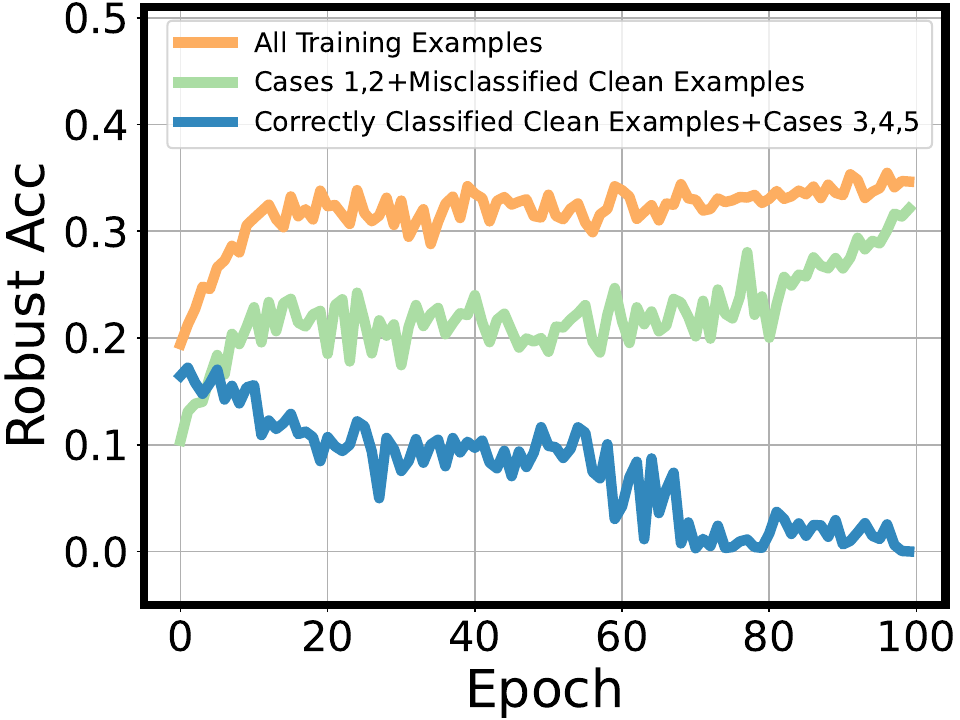}
	\caption{Influences of different case examples on training stability and accuracy. (Left) Clean accuracy. (Right) Robust accuracy.}
	\label{METIImage}
\end{figure}
\subsection{Misclassified Examples and Training Instability}\label{MisInstab}
This part investigates the impact of different examples on training stability. To ensure clarity, we consolidate the five cases as Fig. \ref{TabelSplit} into two situations: correctly classified and misclassified examples. The correctly classified examples include cases 1 and 2. Meanwhile, the misclassified examples include cases 3, 4, and 5. Due to the strong connection between catastrophic overfitting and example taxonomy, we consider two situations. In Fig. \ref{METIImage}, the green line represents only adopting correctly classified clean examples to generate AEs for training, while the remaining misclassified clean examples are directly used for training. The blue line represents only adopting misclassified clean examples to generate AEs, with the other correctly classified examples directly used for training. The orange line indicates that all examples are used to generate AEs for training, which is the standard setup for FAT. Comparing the green and blue lines, it is obvious that the involvement of AEs in misclassified clean examples significantly decreases the robust accuracy and unstabilizes the training. However, simply removing the AEs of misclassified clean examples cannot achieve better performance, as shown in the comparison between green and orange lines. Although the orange line shows instability in training, it has higher robust accuracy than the green line, which indicates that AEs of misclassified clean examples still play important roles in FAT. Therefore, it is essential to effectively utilize misclassified clean examples to enhance the robustness of the model while mitigating the negative impact on catastrophic overfitting.

\begin{algorithm}[t]
	\caption{Select Examples with Different Loss Values}
	\label{SelLoss}
	\textbf{Input}: Example with label $(\mathcal{X}, \mathcal{Y})$; Loss function $\mathcal{L}$; Model $f_\theta(\cdot)$ with weight $\theta$; Learning rate $\sigma$; Loss interval value $\nu$; Total epochs $S$;\\
	\textbf{Return}: Count vector $\vec{\zeta}\in \mathbb{R}^{S\times d}$ with $d$ denotes the number of examples in each loss interval within a training epoch.
	\begin{algorithmic}[1]
	\FOR{$e$ \ in \ $S$}
	\STATE $\vec{\zeta} \gets \vec{0}$;\\
		\FOR{$(\vec{x}, \vec{y})$ \ in \ $(\mathcal{X}, \mathcal{Y})$}
			\STATE $\vec{\delta}_{\text{0}} \gets$ Perturbation initialization;\\
			\STATE $\vec{x_\text{adv}} \gets \vec{x} + \text{Clip}_\epsilon\Big{(}\vec{\delta}_0 + \epsilon \cdot \text{sign}\big{(}\nabla_{\vec{x}}\mathcal{L}(f_{\theta}(\vec{x}+\vec{\delta}_0), \vec{y}\big{)}\Big{)};$\\
			\STATE $l \gets \mathcal{L}\big{(}f_\theta(\vec{x}_\text{adv}),\vec{y}\big{)}$;\\
			\STATE $\vec{\zeta}[e, \lfloor l/\nu \rfloor] += 1$, $\lfloor \cdot \rfloor$ denotes round down;\\
		\ENDFOR
	\ENDFOR
\end{algorithmic}
\end{algorithm}

\subsection{Connection between Loss and Catastrophic Overfitting}\label{CBTL}
\begin{figure}[t]\centering
	\subfigure{\includegraphics[scale=0.265]{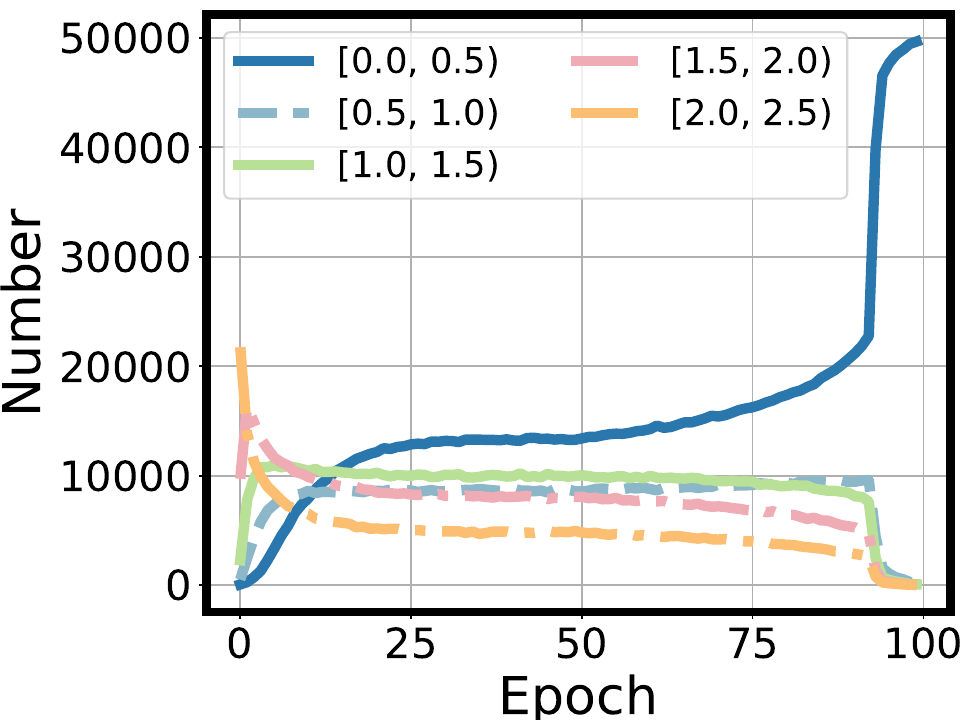}}
	\subfigure{\includegraphics[scale=0.265]{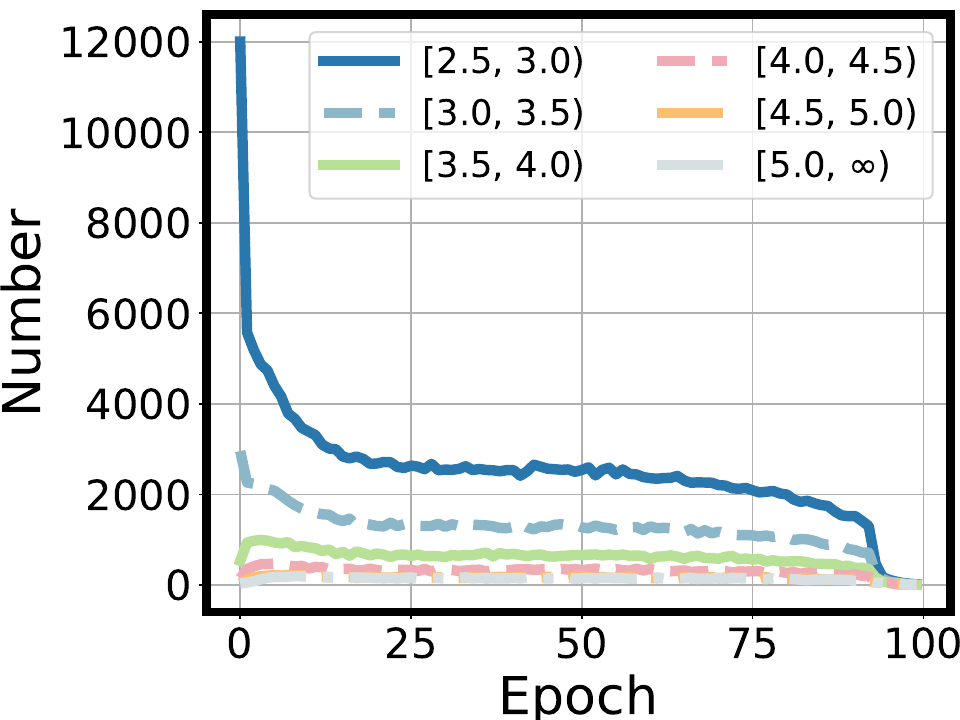}}
	\caption{The change in the number of examples in different loss intervals when performing FGSM-RS \cite{FGSMRS}.}
	\label{RSLC}
\end{figure}

\begin{figure}[t]\centering
	\subfigure{\includegraphics[scale=0.265]{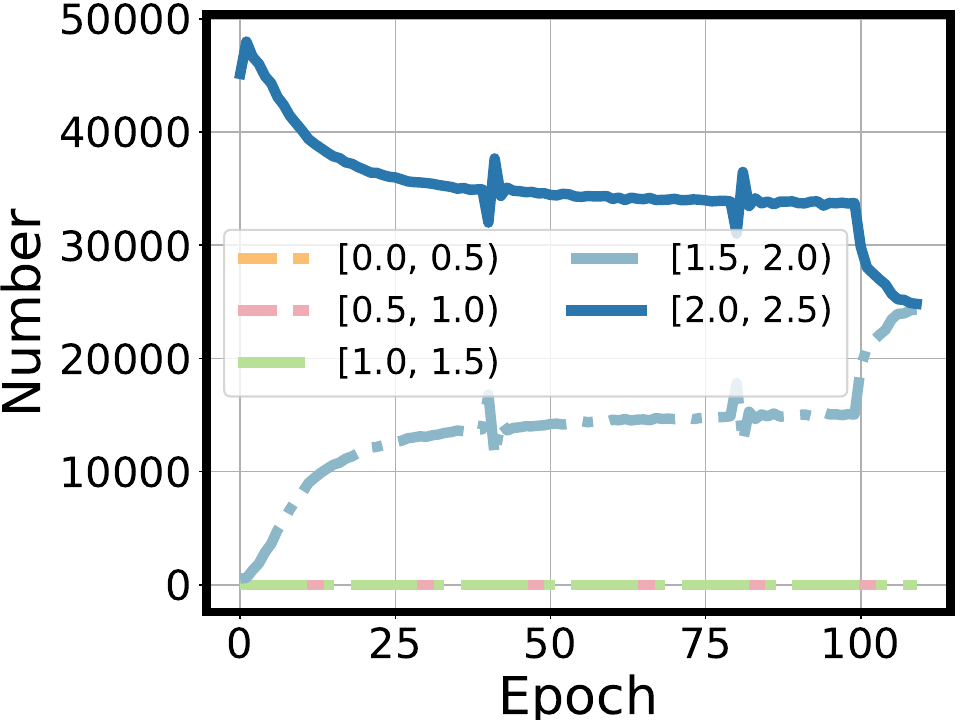}}
	\subfigure{\includegraphics[scale=0.265]{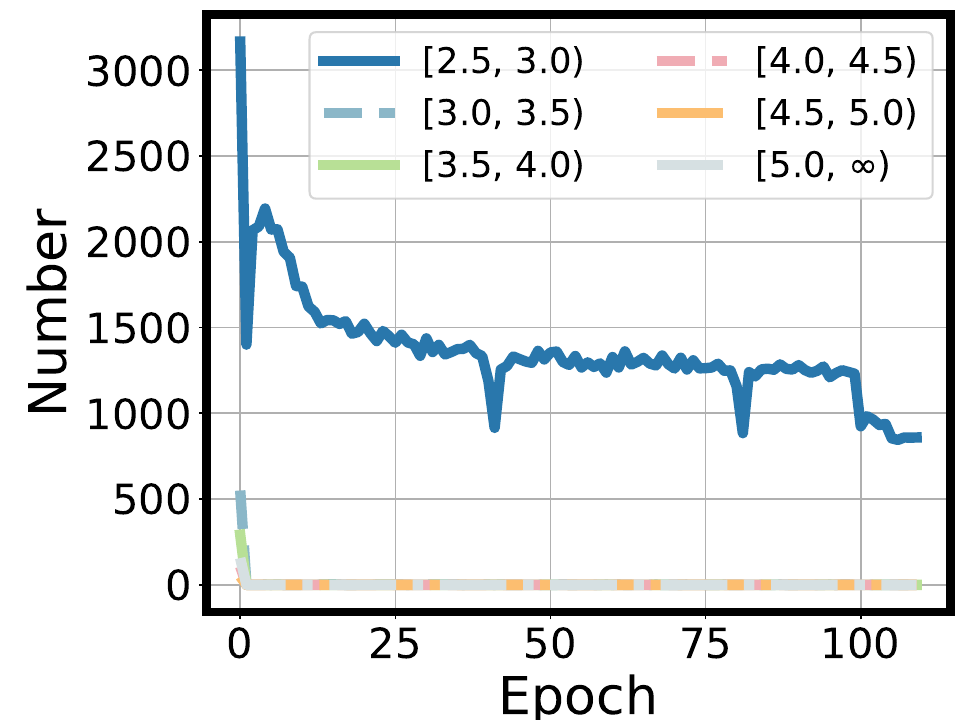}}
	\caption{The change in the number of examples in different loss intervals when performing FGSM-MEP \cite{PGIMEP}.}
	\label{MEPLC}
\end{figure}

As illustrated in previous parts, catastrophic overfitting has a strong connection to the example taxonomy which directly influences the optimization during the training phase. To better explore the variation tendency of AEs during training from all 5 cases mentioned in Section \ref{EPT} as a whole, we propose to investigate the training loss of AEs by dividing them into different loss intervals. We consider two different baselines including FGSM-RS \cite{FGSMRS} and FGSM-MEP \cite{PGIMEP} to empirically demonstrate the connection between training loss of all AEs and catastrophic overfitting which FGSM-RS \cite{FGSMRS} suffers from while FGSM-MEP \cite{PGIMEP} does not. Specifically, we perform training with ResNet18 on CIFAR-10 with a perturbation budget of $8/255$ via FGSM-RS \cite{FGSMRS} and FGSM-MEP \cite{PGIMEP}, as shown in Fig. \ref{RSLC} and \ref{MEPLC} respectively. For each training strategy, we divide all AEs for training into eleven groups based on their loss values and count their number within each loss group. The robust accuracy is evaluated by PGD-10 with $\epsilon=8/255$ and step size $\alpha=2/255$.


Interestingly, an obvious gap between the training losses of these two strategies can be observed. For example, the training losses of AEs with FGSM-RS \cite{FGSMRS} have a dramatic variation. As shown in Fig. \ref{RSLC}, most losses of AEs are within $[1,3.5]$ at the initial training epochs, while the number of examples with loss in range $[0, 0.5)$ gradually increased as training progressed until catastrophic overfitting occurred. Note that these AEs overlap with the situation in case 2 as discussed in subsection \ref{TCO}, indicating that the number of unsuccessfully attacked examples increases. On the contrary, the training loss of AEs with FGSM-MEP \cite{PGIMEP}, which eliminates catastrophic overfitting, has a different tendency. As shown in Fig. \ref{MEPLC}, the training losses of AEs are concentrated in range $[1.5,3.0]$, which is maintained during the entire optimization. We mainly attribute it to the fact that the initialization of perturbation in FGSM-MEP boosts the diversity of AEs since it is close to that of PGD-2 attack \cite{PGIMEP}. With a higher diversity of AEs, the losses can be more concentrated within a range, which alleviates the aforementioned issues. However, FGSM-MEP \cite{PGIMEP} initialization requires simultaneously loading the entire dataset into GPU memory, resulting in significant computational expense when applied to large-scale datasets. Thus, we are motivated to develop an efficient manner to tackle catastrophic overfitting from a different perspective of losses instead of the initialization in FGSM-MEP \cite{PGIMEP}. Based on the observations of the desired concentrated losses, we ask two simple questions:
\begin{itemize}
    \item Does the presence of lower-loss AEs for training negatively impact FAT? 
    \item Could excluding higher-loss AEs help to concentrate the overall losses, thereby eliminating catastrophic overfitting?
\end{itemize}
\begin{table}[t]
	\centering
	\caption{Impact of Discarding AEs in Different Loss Intervals.}
	\begin{tabular}[l]{c| c| c| c c c c c c}
	\toprule[2pt]
	\multirow{1}*{Lower} & \multirow{1}*{Upper} &\multirow{2}*{Number} &\multicolumn{2}{c}{Clean Accuracy} &\multicolumn{2}{c}{Robust Accuracy}\\
	Bound& Bound& &Best& Last &Best&Last \\
	\toprule[1pt]
	0 &10 &$\sim$50000 &83.11 &90.41 &43.25 &1.23  \\
	0 &5 &$\sim$49800 &85.47 &88.71 &44.09 &1.43\\
	0 &4.5 &$\sim$49500 &85.34 &86.77 &44.27 &43.71\\
	0 &4 &$\sim$49000 &85.26 &87.01 &44.66 &44.23\\
	0 &3 &$\sim$31000 &62.20&62.20&33.21&33.21 \\
	0.25 &4 &$\sim$40000 &86.98 &86.98 &44.81 &44.81\\
	0.5 &4 &$\sim$38000 &85.66 &86.98 &44.62 &43.60\\
	0.75 &4 &$\sim$36000 &85.83 &85.83 &41.82 &41.82\\
	1 &4 &$\sim$30000 &84.84 &84.89 &47.00 &45.88\\
	0.5 &$-$ &$\sim$40000 &85.20 &87.56 &43.79 &4.77\\
	1 &$-$ &$\sim$30000 &84.07 &85.18 &45.50 &44.49\\
	1.5 &$-$ &$\sim$25000 &80.18 &80.19 &50.93 & 50.83\\
	2 &$-$ &$\sim$14000 &65.58 &60.95 &45.04 &43.52 \\
	\toprule[2pt]
\end{tabular}
\label{PartAEs}
\end{table}

To address the above questions, we trained ResNet18 using FGSM-RS \cite{FGSMRS}, then analyzed the loss of AEs and discarded AEs within different loss intervals to investigate the impact of various subsets of AEs on the training. The loss partitioning workflow is illustrated in Algorithm \ref{SelLoss}. As shown in the first four rows of Table \ref{PartAEs}, excluding AEs with higher loss can alleviate the catastrophic overfitting and enhance the clean and robust accuracy of the model. This is because excluding examples with high loss can alleviate the over-memorization of the model on these examples, making the training more stable. However, discarding excessive higher-loss examples results in insufficient training data, leading to a noticeable decline in accuracy. In lines 6-9, we analyze the impact of examples with lower-loss on training. By gradually discarding lower-loss examples, we found that catastrophic overfitting does not appear, and both clean and robust accuracy are enhanced. The following two reasons cause this observation. First, AEs with excessively high loss cause the model to over-memorize these few examples, preventing the model learns robust features from the overall data, thereby harming the generalization. Second, AEs with low loss lack adversarial features, and training with these AEs is similar to training with clean examples with random noise, resulting in an inability to learn robust features efficiently. This also explains why catastrophic overfitting occurs, where model adversarial accuracy collapses while clean accuracy remains nearly unchanged. The results from the last four rows show that selecting an appropriate lower limit for the loss can enhance the robustness of the model. However, if the lower limit is too high, deficient training examples lead to decreased clean and robust accuracy. In general, according to Fig. \ref{RSLC} and Table \ref{PartAEs}, AEs with excessively high or low losses are detrimental to training, impairing training stability, and adversarial robustness, respectively. Meanwhile, by narrowing the upper and lower bounds of losses, both clean and robust accuracy can be effectively improved. On the other hand, as observed in Fig. \ref{MEPLC}, FGSM-MEP \cite{PGIMEP} enhances the diversity of AEs, stabilizing and concentrating losses, which can eliminate catastrophic overfitting and improve training performance. 
\par
These findings indicate that concentrating training losses and stabilizing the generation of AEs through increased diversity is crucial for eliminating catastrophic overfitting and improving training performance. At the same time, directly discarding extreme examples may lead to training collapse. This motivates us to develop perturbation initialization methods that enhance AEs diversity and improve the exploitation of training examples based on their loss ranges rather than directly discarding those with negative impacts.

\section{Methodology}\label{Met}
\subsection{Batch Momentum Initialization}
The improvement from FGSM-AT \cite{FGSM} to FGSM-RS \cite{FGSMRS} and FGSM-MEP \cite{PGIMEP} shows the importance of perturbation initialization which can alleviate the catastrophic overfitting and enhance model robustness \cite{FGSMRS, ATSurvey}. The perturbation budget and distribution complexity of the perturbation initialization determine the strength and diversity of AEs \cite{TDAT}. Consequently, we propose a method leveraging historical perturbation information and momentum mechanism to enhance the diversity of AEs without requiring additional calculating time, which is formulated as
\begin{subequations}
	\begin{align}
		&\vec{\delta} = \Phi_{\epsilon}\Big{(}\vec{\delta}_{k-1} + \epsilon \cdot \text{sign}\big{(}\nabla_{\vec{x}}\mathcal{L}(f_{\theta}(\vec{x}+\vec{\delta}_{k-1}), \vec{y})\big{)}\Big{)}, \label{BMIOne} \\
		&\vec{\delta}_{k} = \eta \cdot \vec{\delta}_{k-1} + ( 1 - \eta) \cdot \vec{\delta}, \label{BMITwo}
	\end{align}
\end{subequations}
where $\vec{\delta}_k$ denotes the adversarial perturbation in the $k$-th batch, and $\eta\in[0,1]$ denotes the perturbation momentum trade-off factor to the current and historical perturbation. The perturbation $\vec{\delta}_{k-1}$ is sampled from a uniform distribution in the first training batch. Note that our batch momentum initialization does not require additional computational cost, which only requires storing additional data related to the training batch size. This method effectively enhances the diversity of AEs by introducing the momentum mechanism, making the generation of AEs more stable. Consequently, it reduces the number of lower-loss and higher-loss AEs, concentrating the training losses. Therefore, batch momentum initialization can stabilize training and mitigate catastrophic overfitting.

\subsection{Dynamic Label Relaxation}\label{DLRDefSec}
As analyzed in subsection \ref{MisInstab}, the AEs generated by misclassified clean examples in FAT could significantly influence the imbalance in min-max optimization, however, directly discarding them cannot lead to better performance. Together with the observation in subsection \ref{CBTL} that the loss range is expected to be concentrated, we propose to alleviate the higher and lower losses during FAT through the involvement of a dynamic label relaxation. Formally, the dynamic label relaxation can be formulated as
\begin{equation}\label{DLRDef}
	\vec{\hat{y}} = \vec{y}\cdot\gamma + \left( \vec{y} - \vec{1} \right) \cdot \frac{\gamma-1}{m-1},
\end{equation}
where $\vec{\hat{y}}$ denotes the relaxation label, $\vec{y}$ signifies one-hot label, $m$ represents the number of classes for the training, and $\gamma$ represents the label relaxation factor, which is formulated as
\begin{equation}\label{DLRDetail}
	\gamma = \begin{cases}
		\begin{aligned}
			&\beta\cdot\tanh(1-\Delta), \text{if}\ \beta\cdot\tanh(1-\Delta) \geq \gamma_{\text{min}},\\
			&\gamma_{\text{min}}, \quad\quad\quad\quad\quad \text{if}\ \beta\cdot\tanh(1-\Delta) < \gamma_{\text{min}},
		\end{aligned}
	\end{cases}
\end{equation}
where $\Delta = e/S$ with $e$ and $S$ denote the current and total training epochs, respectively. The hyperparameter $\beta>0$ controls the amplitude of relaxation, while the parameter $\gamma$ gradually decreases with training. The parameter $\gamma_\text{min}$ ensures the relaxed label correctly guides the updates of the model. Meanwhile, the function $\tanh(\cdot)$ is furnished to accelerate the reduction of $\gamma$. As analyzed in subsection \ref{TCO}, the imbalance optimization in FAT results in catastrophic overfitting. Our dynamic label relaxation adjusts the label relaxation magnitude according to different training stages, thereby stabilizing the training. The parameter $\gamma$ of the dynamic label relaxation decreases as training progresses and is reduced to a predefined minimum value. During the initial epochs, this method uses one-hot labels to assist the model in capturing the primary feature distribution of the examples, thereby rapidly optimizing the accuracy. As training advances, this method increases the label smoothness, encouraging the model to classify examples correctly without pursuing excessively high confidence, which contributes to the concentration of loss range. This helps the network improve accuracy on both clean examples and AEs.

\subsection{Taxonomy Driven Loss}
Inspired by the observation of misclassified examples and catastrophic overfitting presented in subsection \ref{TCO}, we propose a taxonomy driven loss with regularization to mitigate the instability caused by misclassified examples. First, we review the definition of standard cross-entropy loss as
\begin{equation}\label{CEDef}
	\mathcal{L}_{\text{CE}}=-\sum^{m}_{d=1} \vec{\hat{y}}_d \log f_\theta(\vec{x} + \vec{\delta})_d,
\end{equation}
where $\vec{\hat{y}}_d$ represents the $d$-th subelement of $\vec{\hat{y}}$ and $f_\theta(\cdot)_d$ represents the $d$-th subelement of model output. To mitigate the impact of AEs generated from misclassified clean examples on training, the taxonomy driven loss function is established as
\begin{equation}\label{TDLDef}
	\mathcal{L}_{\text{TD}}=\mathcal{L}_{\text{CE}}+\lambda \cdot \|f(\vec{x}+\vec{\delta})-f(\vec{x})\|_2 \cdot \tanh(1-p),
\end{equation}
where $p$ denotes the confidence of groundturth class and hyperparameter $\lambda>0$. In the case of $p=1$ the $\mathcal{L}_{\text{TD}}$ is equivalent to cross entropy loss $\mathcal{L}_{\text{CE}}$. As observed in section \ref{ProblemDes}, the AEs generated by the misclassified clean examples negatively impact training stability. Introducing the amplification term $\tanh(1-p)$ aims to more rigorously separate and penalize low-confidence training examples, which are often misclassified by the model. According to the analysis in subsection \ref{MisInstab}, adopting misclassified training examples negatively impacts the stability of training process. To this end, taxonomy driven loss improves the classification ability of the misclassified examples while maintaining the existing classification performance of the model. This objective is achieved by assigning smaller losses to correctly classified examples and larger losses to incorrectly classified examples. Thus, this approach can minimize the number of misclassified examples, thereby enhancing training effectiveness and stability.

\begin{algorithm}[t]
	\caption{ETA}
	\label{AlgTAEE}
	\textbf{Parameters}: Training data $\mathcal{X}$ and label $\mathcal{Y}$; Learning rate $\mu$; Number of training epochs $S$; Relaxation factor $\gamma$; Momentum factor $\eta$; scale factor $\beta$;\\
	\textbf{Return}: Robust model $f_\theta(\cdot)$;
    \begin{algorithmic}[1] 
	\STATE $\vec{\delta}_{\text{0}} \gets \mathcal{U}(-\epsilon, \epsilon)$.
	\FOR{$e$ in $S$}
        \STATE $\gamma\gets\beta\cdot\text{tanh}(1-\frac{e}{S})$;\\
		\IF {$\gamma < \gamma_{\text{min}}$}
			\STATE $\gamma \gets \gamma_{\text{min}}$;
		\ENDIF
		\STATE $k \gets 1$;\\
		\FOR{$(\vec{x},\vec{y})$ in $(\mathcal{X},\mathcal{Y})$}
            \STATE $p \gets \text{Softmax}(f(\vec{x})) [d]  \text{, where } \vec{y}[d] = 1$;\\
			\STATE $\hat{\vec{y}} \gets \vec{y} \cdot \gamma + \left(\vec{y} - \vec{1} \right) \cdot \frac{\gamma-1}{m-1}$;\\
			\STATE $\vec{\delta} \gets \text{Clip}_\epsilon\Big{(} \vec{\delta}_{k-1}+\epsilon\cdot \text{sign}\big{(}\nabla_{\vec{x}}\mathcal{L}(f_\theta(\vec{x}+\vec{\delta}_{k-1}),\vec{\hat{y}})\big{)}\Big{)}$;\\
			\STATE $\mathcal{L}_{\text{TD}}\gets\mathcal{L}_{\text{CE}}+\lambda \cdot \|f(\vec{x}+\vec{\delta})-f(\vec{x})\|_2 \cdot \tanh(1-p);$\\
			\STATE $\Omega=[\eta~\text{if}~f_\theta(\vec{x})=\vec{y}~\text{else}~1];$\\
			\STATE $\Psi = [l_1, l_2, \cdots, l_i, \cdots, l_m];$\\
			\STATE $\theta\gets\theta-\mu \nabla_\theta \mathcal{L}_{\text{CG}}$;\\
			\STATE $\vec{\delta}_{k} \gets \eta \cdot \vec{\eta}_{k-1} + (1-\eta) \cdot \vec{\delta}$;\\
			\STATE $k \gets k+1$.
		\ENDFOR
	\ENDFOR
\end{algorithmic}
\end{algorithm}

\subsection{Catastrophic Overfitting Aware Loss Adaptation}\label{COALA}
Based on the analysis in subsection \ref{CBTL}, both excessively large and small losses of AEs used for FAT are detrimental to performance. Therefore, keeping the example losses within an appropriate range is essential. Our experiments show that directly discarding a portion of training AEs with small losses in the FGSM-RS \cite{FGSMRS} can improve clean and robust accuracy. However, experiments with the FGSM-MEP \cite{PGIMEP} indicate that most losses are relatively concentrated, and directly discarding AEs could harm the performance of models. To exploit this observation for improvements and ensure the scalability of our method, the COLA is presented, which adjusts the loss effect of different examples instead of directly discarding them.
\par
Define clean training set $X=[\vec{x}_1, \cdots, \vec{x}_a \cdots, \vec{x}_n]$ with $n$ examples and their labels as $Y=[\vec{y}_1, \cdots, \vec{y}_a,\cdots, \vec{y}_n]$. Then, performing adversarial attack to generate the AEs for training as $X_{\text{adv}}=[\vec{x}^1_{\text{adv}}, \cdots, \vec{x}^a_{\text{adv}} \cdots, \vec{x}^n_{\text{adv}}]$, where the loss set vector $\Psi\in\mathbb{R}^{n}$ of these AEs is defined as
\begin{equation}
	\Psi = \mathcal{L}(f_\theta(X_{\text{adv}}),Y) = [l_1, l_2, \cdots, l_a, \cdots, l_n]^{\text{T}},
\end{equation}
where $^{\text{T}}$ denotes the transpose operation. After that, a loss selector $\Omega\in\mathbb{R}^{n}$ is constructed. For each element in loss selector $\Omega$, if the prediction result of model $f_\theta(\cdot)$ equals the groundtruth label, then the value of the $a$-th element of $\Omega$ is the loss adaptation factor $\eta_c\in (0,1)$. Meanwhile, in case of $f_\theta(\vec{x}_a)\ne \vec{y}_a$, the corresponding position in $\Omega$ is set as $1$. Thereafter, the loss for each AE for training is adjusted as
\begin{equation}\label{LCGDef}
	\mathcal{L}_{\text{CG}} = \Psi^\text{T}\Omega.
\end{equation}
The following training is consistent with the standard FAT procedure, using $\mathcal{L}_{\text{CG}}$ for backpropagation to update the model.

\vspace{-0.3cm}
\subsection{Method Implementation}
By integrating the proposed batch momentum initialization, dynamic label relaxation, taxonomy driven loss, and COLA, our adversarial training paradigm is concluded, dubbed ETA. The pseudocode description of ETA is presented in Algorithm \ref{AlgTAEE}. First, the initialization perturbation $\vec{\delta}$ is sampled from a uniform distribution $\mathcal{U}(-\epsilon, \epsilon)$. During training, each batch initializes the perturbation according to batch momentum initialization \eqref{BMIOne}--\eqref{BMITwo} and configures the labels using dynamic label relaxation \eqref{DLRDef}. After that, AEs with initialization perturbation and relaxation label is generated by the FGSM. Subsequently, the training loss is calculated using taxonomy driven loss \eqref{TDLDef} and optimized with the COLA \eqref{LCGDef}. Finally, the loss is used for backpropagation to update the model $f_\theta(\cdot)$. Our ETA systematically enhances the FAT, encompassing perturbation initialization, loss adaptation, and training loss.

\begin{table*}[t]
	\centering
	\caption{Comparison Results on CIFAR-10/100 Datasets. Bold Font Indicates the Best Result}
	\setlength{\tabcolsep}{1.0mm}{
	\begin{tabular}{l| c| c c c c c c c c| c c c c c c c c}
	\toprule[2pt]
	\multirow{2}*{Method} & &\multicolumn{8}{c|}{CIFAR-10} &\multicolumn{8}{c}{CIFAR-100}\\
	&&\multirow{1}*{Clean} &\multirow{1}*{FGSM} &\multirow{1}*{MI} &\multirow{1}*{PGD10} &\multirow{1}*{PGD20} &\multirow{1}*{PGD50} &\multirow{1}*{AA} &\multirow{1}*{CW} &\multirow{1}*{Clean} &\multirow{1}*{FGSM} &\multirow{1}*{MI} &\multirow{1}*{PGD10} &\multirow{1}*{PGD20} &\multirow{1}*{PGD50} &\multirow{1}*{AA} &\multirow{1}*{CW}\\
	\toprule[1pt]
	\multicolumn{18}{c}{\qquad \qquad \qquad \qquad \qquad Fast Adversarial Training}\\
	\toprule[1pt]
	\multirow{2}*{FGSM-RS\cite{FGSMRS}} &Best &\textbf{83.69} &62.00 &48.95 &47.66 &46.29 &45.96 &42.80 &46.10 &51.67 &31.02 &23.05 &22.61 &22.04 &21.75 &18.72 &20.92\\
	&Last &83.69 &62.00 &48.95 &47.66 &46.29 &45.96 &42.80 &46.10 &51.67 &31.02 &23.05 &22.61 &22.04 &21.75 &18.72 &20.92\\
	\toprule[0.5pt]
	\multirow{2}*{Free-AT(m=8)\cite{FreeAT}} &Best &81.38 &60.81 &49.92 &49.07 &48.03 &47.62 &44.37 &46.98 &52.06 &32.13 &25.06 &24.74 &24.09 &24.04 &20.23 &22.43\\
	&Last &81.38 &60.81 &49.92 &49.07 &48.03 &47.62 &44.37 &46.98 &52.06 &32.13 &25.06 &24.74 &24.09 &24.04 &20.23 &22.43\\
	\toprule[0.5pt]
	\multirow{2}*{GAT\cite{GAT}} &Best &81.53 &64.18 &54.88 &54.05 &53.26 &52.95 &47.68 &49.76 &57.49 &36.77 &29.63 &29.14 &28.60 &28.30 &23.11 &25.14\\
	&Last &81.88 &64.30 &53.89 &53.23 &52.16 &51.86 &47.08 &49.71 &57.58 &36.85 &29.55 &29.06 &28.43 &28.30 &23.02 &24.97\\
	\toprule[0.5pt]
	\multirow{2}*{FGSM-SDI\cite{SDI}} &Best &83.55 &63.60 &52.75 &51.94 &50.65 &50.34 &46.31 &49.09 &58.64 &37.23 &29.19 &28.78 &27.99 &27.67 &23.27 &25.85\\
	&Last &\textbf{83.73} &63.75 &52.77 &51.88 &50.49 &50.09 &46.34 &49.42 &58.54 &37.19 &29.17 &28.71 &28.00 &27.72 &23.18 &25.55\\
	\toprule[0.5pt]
	\multirow{2}*{FGSM-MEP\cite{PGIMEP}} &Best &81.71 &65.02 &55.48 &55.26 &54.54 &54.38 &48.20 &\textbf{50.72} &\textbf{58.78} &40.02 &31.84 &31.94 &31.30 &31.19 &25.65 &28.23\\
	&Last &81.71 &65.02 &55.48 &55.26 &54.54 &54.38 &48.20 &\textbf{50.72} &\textbf{58.82} &39.83 &31.56 &31.65 &31.18 &30.89 &25.43 &27.75\\
	\toprule[0.5pt]
	\multirow{2}*{GradAlign\cite{GradAlign}} &Best &80.45 &60.56 &50.00 &49.11 &47.96 &47.63 &43.92 &46.94 &54.90 &35.28 &27.50 &27.13 &26.52 &26.22 &22.30 &25.01\\
	&Last &80.45 &60.56 &50.00 &49.11 &47.96 &47.63 &43.92 &46.94 &55.22 &35.51 &27.40 &27.12 &26.42 &26.24 &22.19 &24.94\\
	\toprule[0.5pt]
	\multirow{2}*{NFGSM\cite{NFGSM}} &Best &80.35 &60.93 &50.80 &49.83 &48.77 &48.51 &44.54 &47.37 &54.41 &35.00 &27.59 &27.01 &26.55 &26.34 &22.81 &25.08\\
	&Last &80.35 &60.93 &50.80 &49.83 &48.77 &48.51 &44.54 &47.37 &54.41 &35.00 &27.59 &27.01 &26.55 &26.34 &22.81 &25.08\\
	\toprule[0.5pt]
	\multirow{2}*{TDAT\cite{TDAT}} &Best &82.25 &66.15 &57.40 &56.85 &56.05 &55.90 &48.33 &50.18 &57.32 &40.29 &33.73 &33.56 &33.17 &33.06 &\textbf{26.61} &28.47\\
	&Last &82.25 &66.15 &57.40 &56.85 &56.05 &55.90 &48.33 &50.18 &57.32 &40.29 &33.73 &33.56 &33.17 &33.06 &\textbf{26.61} &28.47\\
	\toprule[0.5pt]
	\rowcolor{black!10}\multirow{2}*{ETA (Ours)} &Best &83.19 &\textbf{66.98} &\textbf{57.89} &\textbf{57.63} &\textbf{56.83} &\textbf{56.63} &\textbf{48.66} &49.86 &57.54 &\textbf{41.33} &\textbf{34.33} &\textbf{34.33} &\textbf{33.83} &\textbf{33.80} &26.36 &\textbf{28.83}\\
	\rowcolor{black!10}\multirow{-2}*{ETA (Ours)}&Last &83.21 &\textbf{66.69} &\textbf{57.66} &\textbf{57.31} &\textbf{56.41} &\textbf{56.22} &\textbf{48.42} &50.08 &57.40 &\textbf{40.85} &\textbf{33.99} &\textbf{34.14} &\textbf{33.66} &\textbf{33.51} &26.05 &\textbf{28.71}\\
	\toprule[1pt]
	\multicolumn{18}{c}{\qquad \qquad \qquad \qquad \qquad Multi-step Adversarial Training}\\
	\toprule[1pt]
	\multirow{2}*{MART\cite{MART}} &Best &82.03 &64.94 &55.64 &54.83 &53.72 &53.53 &47.74 &49.68 &54.51 &38.62 &32.37 &32.18 &31.68 &31.59 &26.07 &28.01\\
	&Last &82.33 &65.12 &55.21 &54.38 &52.98 &52.60 &47.46 &49.66 &54.75 &38.52 &32.18 &31.85 &31.37 &31.21 &25.71 &27.81\\
	\toprule[0.5pt]
	\multirow{2}*{LAS-AWP\cite{LASAT}} &Best &82.92 &65.86 &57.05 &56.37 &55.57 &55.20 &49.46 &51.53 &58.75 &40.66 &32.98 &32.58 &31.91 &31.74 &27.23 &29.59\\
	&Last &82.92 &65.86 &57.05 &56.37 &55.57 &55.20 &49.46 &51.53 &58.75 &40.66 &32.98 &32.58 &31.91 &31.74 &27.23 &29.59\\
	\toprule[2pt]
    \end{tabular}}
    \label{CIFARResults}
\end{table*}

\begin{table*}[t]
	\centering
	\caption{Comparison Results on Tiny ImageNet and ImageNet-100 Datasets. Bold Font Indicates the Best Result}
	\setlength{\tabcolsep}{1.0mm}{
	\begin{tabular}[l]{l| c| c c c c c c c c| c c c c c c c c}
	\toprule[2pt]
	\multirow{2}*{Method} & &\multicolumn{8}{c|}{Tiny ImageNet} &\multicolumn{8}{c}{ImageNet-100}\\
	&&\multirow{1}*{Clean} &\multirow{1}*{FGSM} &\multirow{1}*{MI} &\multirow{1}*{PGD10} &\multirow{1}*{PGD20} &\multirow{1}*{PGD50} &\multirow{1}*{AA} &\multirow{1}*{CW} &\multirow{1}*{Clean} &\multirow{1}*{FGSM} &\multirow{1}*{MI} &\multirow{1}*{PGD10} &\multirow{1}*{PGD20} &\multirow{1}*{PGD50} &\multirow{1}*{AA} &\multirow{1}*{CW}\\
	\toprule[1pt]
	\multicolumn{18}{c}{\qquad \qquad \qquad \qquad \qquad \qquad Fast Adversarial Training}\\
	\toprule[1pt]
	\multirow{2}*{FGSM-RS\cite{FGSMRS}} &Best &43.52 &23.93 &17.60 &17.22 &16.82 &16.64 &13.09 &14.67 &54.62 &36.24 &22.60 &26.60 &25.64 &25.34 &19.30 &23.74\\
	&Last &43.52 &23.93 &17.60 &17.22 &16.82 &16.64 &13.09 &14.67 &54.62 &36.24 &22.60 &26.60 &25.64 &25.34 &19.30 &23.74\\
	\toprule[0.5pt]
	\multirow{2}*{Free-AT(m=8)\cite{FreeAT}} &Best &44.15 &25.18 &18.36 &17.95 &17.47 &17.31 &13.67 &15.82 &51.48 &32.24 &23.72 &23.64 &22.92 &22.80 &18.86 &21.38\\
	&Last &44.15 &25.18 &18.36 &17.95 &17.47 &17.31 &13.67 &15.82 &51.48 &32.24 &23.72 &23.64 &22.92 &22.80 &18.86 &21.38\\
	\toprule[0.5pt]
	\multirow{2}*{GAT\cite{GAT}} &Best &46.00 &23.04 &15.63 &15.16 &14.51 &14.33 &10.82 &13.27 &\textbf{66.42} &47.74 &37.62 &39.16 &38.08 &37.62 &29.04 &32.90\\
	&Last &45.57 &22.10 &15.03 &14.56 &14.03 &13.85 &10.26 &12.71 &\textbf{66.42} &47.74 &37.62 &39.16 &38.08 &37.62 &29.04 &32.90\\
	\toprule[0.5pt]
	\multirow{2}*{FGSM-SDI\cite{SDI}} &Best &43.71 &26.84 &20.80 &20.60 &20.26 &20.11 &15.43 &17.16 &62.42 &47.12 &38.60 &39.00 &38.42 &38.28 &\textbf{30.38} &33.40\\
	&Last &45.40 &24.76 &17.60 &17.27 &16.84 &16.73 &12.47 &14.80 &62.42 &47.12 &38.60 &39.00 &38.42 &38.28 &\textbf{30.38} &33.40\\
	\toprule[0.5pt]
	\multirow{2}*{FGSM-MEP\cite{PGIMEP}} &Best &42.98 &28.55 &23.43 &23.27 &23.01 &22.92 &17.00 &18.67 &63.24 &48.22 &36.88 &39.26 &38.70 &38.38 &29.90 &33.61\\
	&Last &45.13 &28.11 &21.78 &21.51 &21.19 &21.07 &14.86 &16.86 &63.52 &48.54 &36.46 &38.98 &38.26 &38.04 &29.38 &33.26\\
	\toprule[0.5pt]
	\multirow{2}*{GradAlign\cite{GradAlign}} &Best &38.22 &22.85 &17.35 &17.20 &16.86 &16.79 &12.64 &14.00 &- &- &- &- &- &- &- &- \\
	&Last &37.89 &22.51 &17.22 &17.06 &16.78 &16.69 &12.49 &13.90&- &- &- &- &- &- &- &- \\
	\toprule[0.5pt]
	\multirow{2}*{NFGSM\cite{NFGSM}} &Best &\textbf{46.06} &24.72 &17.25 &16.74 &16.21 &16.02 &12.71 &14.96 &61.52 &43.58 &35.36 &34.64 &33.86 &33.52 &27.76 &30.98\\
	&Last &\textbf{46.06} &24.72 &17.25 &16.74 &16.21 &16.02 &12.71 &14.96 &61.52 &43.58 &35.36 &34.64 &33.86 &33.52 &27.76 &30.98\\
	\toprule[0.5pt]
	\multirow{2}*{TDAT\cite{TDAT}} &Best &43.60 &\textbf{30.38} &24.48 &23.98 &23.61 &23.50 &16.17 &18.44 &63.14 &48.36 &38.62  &40.52 &39.92 &39.78 &29.80 &33.44\\
	&Last &43.67 &\textbf{30.31} &24.26 &23.86 &23.54 &23.40 &16.05 &18.31 &63.76 &48.68 &38.08 &39.92 &39.34 &39.08 &30.00 &\textbf{33.56}\\
	\toprule[0.5pt]
	\rowcolor{black!10}\multirow{2}*{ETA (Ours)} &Best &41.67 &30.12 &\textbf{25.76} &\textbf{25.21} &\textbf{24.96} &\textbf{24.92} &\textbf{17.20} &\textbf{19.43} &63.90 &\textbf{48.84} &\textbf{38.76} &\textbf{40.60} &\textbf{40.08} &\textbf{39.86} &29.68 &\textbf{33.72}\\
	\rowcolor{black!10}\multirow{-2}*{ETA (Ours)}&Last &41.95 &30.06 &\textbf{25.45} &\textbf{24.94} &\textbf{24.71} &\textbf{24.56} &\textbf{16.69} &\textbf{19.14} &63.88 &\textbf{48.84} &\textbf{38.70} &\textbf{40.42} &\textbf{39.84} &\textbf{39.54} &29.64 &33.45\\
	\toprule[1pt]
	\multicolumn{18}{c}{\qquad \qquad \qquad \qquad \qquad Multi-step Adversarial Training}\\
	\toprule[1pt]
	\multirow{2}*{MART\cite{MART}} &Best &38.41 &25.20 &21.15 &20.93 &20.74 &20.67 &15.53 &16.88 &64.80 &47.88 &39.94  &39.28 &38.34 &37.96 &31.86 &35.12\\
	&Last &36.83 &18.02 &12.70 &12.36 &12.01 &11.93 &9.26 &10.36 &64.80 &47.88 &39.94 &39.28 &38.34 &37.96 &31.86 &35.12\\
	\toprule[0.5pt]
	\multirow{2}*{LAS-AWP\cite{LASAT}} &Best &47.86 &30.77 &24.53 &24.10 &23.67 &23.60 &18.21 &20.49 &64.52 &48.80 &38.84 &40.62 &40.24 &40.18 &32.46 &35.46\\
	&Last &47.86 &30.77 &24.53 &24.10 &23.67 &23.60 &18.21 &20.49 &64.54 &48.80 &38.76 &40.62 &40.18 &40.08 &32.26 &35.38\\
	\toprule[2pt]
    \end{tabular}}
    \label{ImageNetResults}
\end{table*}

\section{Experiments and Analysis}\label{EandA}

\subsection{Experimental Setup}
\subsubsection{Experiments Environment}
Our hardware environment adopts Intel(R) Xeon(R) Silver 4314 CPU @ 2.40GHz and NVIDIA GeForce RTX 4090 GPU with 24GB memory. Due to limited computational resources, the computational complexity comparison is executed using the single RTX 3090 GPU. 

\subsubsection{Datasets and Training Settings}
The experiments are executed in four datasets, covering CIFAR-10/100 \cite{cifar}, Tiny ImageNet \cite{tinyimagenet}, and ImageNet-100 \cite{imagenet}. To perform the experiments, we adopt ResNet18 \cite{ResNet}. We adopt the SGD optimizer with a momentum of 0.9 and weight decay of 5e-4. On the CIFAR-10/100 and Tiny ImageNet, training is conducted for 110 epochs. The initial learning rate is 0.1 and is reduced by a factor of 10 at the 100 and 105 epochs. For the ImageNet-100 dataset, training is performed for 50 epochs, with the initial learning rate set to 0.1 and divided by 10 at the 40 and 45 epochs. The batch size for all experiments is set to 128 \cite{PGK}. 

\subsubsection{Attacks for Evaluation}
We selected representative adversarial attacks for a comprehensive evaluation, including FGSM \cite{FGSM}, MIFGSM (MI) \cite{MIFGSM}, PGD-10/20/50 \cite{PGD}, AutoAttack (AA) \cite{AutoAttack}, and C$\&$W (CW) \cite{CW}. The attack budget $\epsilon=8/255$ with step size for multi-step attacks is set as $\alpha=2/255$. We follow the implementations in TorchAttack and employ the default settings \cite{torchattacks}.

\subsubsection{Baselines}
We consider representative and state-of-the-art methods for comparison. Specifically, FAT for comparison include FGSM-RS \cite{FGSMRS}, Free-AT \cite{FreeAT}, GAT \cite{GAT}, FGSM-SDI \cite{SDI}, FGSM-MEP \cite{PGIMEP}, GradAlign \cite{GradAlign}, NFGSM \cite{NFGSM}, and TDAT \cite{TDAT}. For multi-step adversarial training, we include MART \cite{MART} and LAS-AWP \cite{LASAWP}.

\subsubsection{Last and Best Epoch Defination}
Reporting the best and last epoch checkpoint performance allows for a comprehensive evaluation of best performance and stability. The ``best'' checkpoint represents the optimal robustness of the model against PGD during training, while the ``last'' checkpoint represents the robustness of the model after training. After each epoch, the model is evaluated using PGD-10. For example, if the model obtained its best performance against PGD at the 90th epoch, then the ``best'' checkpoint comes from this epoch. Meanwhile, the ``last'' checkpoint comes from the model after finishing all epochs. After training, the performance of the best and last checkpoints is tested under various adversarial attacks. Given that the selection for the best checkpoint relies entirely on PGD-10, the performance of the ``last'' checkpoint could be better than that of the ``best'' checkpoint when against other adversarial attacks \cite{TDAT, PGIMEP}.

\subsection{Comparison Experiments and Analysis}\label{CEA}
\subsubsection{Results on CIFAR-10} 
The comparison results on CIFAR-10 are shown in Table \ref{CIFARResults}, with the best results highlighted in bold. Our ETA achieves state-of-the-art robustness against FGSM, MIFGSM, PGD, and AA while maintaining competitive clean accuracy. The results reveal that our ETA can avoid catastrophic overfitting and achieves robust accuracy improvements. Except for TDAT, the proposed ETA achieves improvement over existing FAT methods, which achieves the best robustness at the best model checkpoint against FGSM (+1.96$\%$), MIFGSM (+2.41$\%$), PGD-10 (+2.37$\%$), PGD-20 (+2.29$\%$), PGD-50 (+2.25$\%$), and AA (+0.33$\%$). Compared with TDAT, our ETA achieves better clean and robust accuracy. Additionally, the results confirm that ETA demonstrates better robustness than multi-step adversarial training.

\subsubsection{Results on CIFAR-100}
The comparison results on CIFAR-100 are shown in Table \ref{CIFARResults}. Our ETA exhibits competitive performance compared to FAT and multi-step adversarial training, achieving the best robustness against five attack methods. Specifically, Our ETA achieves state-of-the-art robust accuracy against FGSM, MIFGSM, PGD, APGD, and CW attacks while also demonstrating competitive clean accuracy. On the other aspect, the ETA achieves the best robust accuracy in the last checkpoint, demonstrating training stability. Therefore, avoids the requirement for early stopping and reduces the resources required for training by not having to compare the performance of each checkpoint.

\subsubsection{Results on Tiny ImageNet}
The comparison of the results on Tiny ImageNet is shown in Table \ref{ImageNetResults}. The Tiny ImageNet contains more data and classes than CIFAR-10/100, thereby making it more challenging to maintain training stability and improve robust accuracy. Specifically, our proposed ETA achieves state-of-the-art results against MIFGSM (+1.28$\%$), PGD-10 (+1.23$\%$), PGD-20 (+1.35$\%$), PGD-50 (+1.42$\%$), AA (+1.03$\%$), and C$\&$W (+0.99$\%$) at the best model checkpoint. Meanwhile, the robust accuracy generated by the ETA on the Tiny ImageNet possesses the overall best performance with only a slight gap between the best and last epoch. 

\subsubsection{Results on ImageNet-100}
The comparison of the results on ImageNet-100 is shown in Table \ref{ImageNetResults}. Compared to other methods except TDAT, our ETA demonstrates competitive robustness accuracy at the best checkpoint against FGSM (+0.62$\%$), MIFGSM (+0.16$\%$), PGD-10 (+1.34$\%$), PGD-20 (+1.38$\%$), and PGD-50 (+1.48$\%$). Compared to the TDAT, ETA improves both clean and robustness accuracy when against FGSM, MIFGSM, PGD, and CW attacks. Specifically, the last checkpoint generated by the ETA achieves better performance in clean accuracy (+0.12$\%$), FGSM (+0.18$\%$), MIFGSM (+0.62$\%$), PGD-10 (+0.5$\%$), PGD-20 (+0.5$\%$), and PGD-50 (+0.46$\%$). Experimental results demonstrate that ETA is more advantageous on large and complex datasets, achieving the best robustness accuracy against different adversarial attacks while maintaining competitive clean accuracy.

\subsection{Combining with Existing Methods}\label{CwithEM}
COLA can be a plugin to enhance other methods. The experimental results are shown in Table \ref{pluginResults}, with bold indicating the improved results when combined with the loss adjustment method. We selected five FAT methods, covering FGSM-RS \cite{FGSMRS}, NFGSM \cite{NFGSM}, GAT \cite{GAT}, FGSM-MEP \cite{PGIMEP}, and TDAT \cite{TDAT} as baselines. These methods are trained on the CIFAR-10 dataset using ResNet18, with a loss adjustment factor of 0.5. Following standard experimental protocols, we evaluate robustness at the last and best checkpoints. Table \ref{pluginResults} presents the results for each method with and without loss adjustment. The experimental results indicate that COLA can be efficiently plugged into other methods. This is achieved by determining whether the AEs can be correctly classified and adjusting the loss accordingly. With the involvement of COLA, all the baselines achieve better resistance against gradient-based attacks when the same loss adjustment factor is chosen. Specifically, the average robust accuracy of the best checkpoint is improved by $0.31\%$ for FGSM-RS, $1.65\%$ for NFGSM, $1.37\%$ for GAT, $1.69\%$ for FGSM-MEP, and $2.18\%$ for TDAT. Adversarial training faces a trade-off between clean and robust accuracy. Enhancing robustness usually compromises clean accuracy. However, the proposed loss adjustment method can increase model robustness with minimal reduction in clean accuracy. For instance, with the involvement of the proposed loss adjustment, the NFGSM achieves better trade-offs, which maintains clean accuracy while improving robustness. Similar improvements can be observed in FGSM-MEP and TDAT, which show improvements in robust and clean accuracy.

\begin{table}[t]
	\centering
	\caption{Results of Combining COLA with Existing Methods. AR Represents the Average Robust Accuracy. Bold Font Indicates Results That are Better Than the Original Method.}
	\setlength{\tabcolsep}{0.95mm}{
	\begin{tabular}[l]{c| c| c| c c c c c c}
	\toprule[2pt]
	Method & &Clean &FGSM &MI &PGD10 &PGD20 &PGD50 &AR\\
	\toprule[1pt]
	\multirow{2}*{FGSM-RS} &Best &86.75 &62.19 &46.41 &44.46 &42.38 &41.68 &47.42\\
	&Last &87.89 &62.77 &45.83 &43.77 &41.54 &40.96 &46.97\\
	\cline{2-9}
	\rowcolor{black!10}\multirow{2}*{+ COLA}&Best &86.71 &\textbf{62.71} &46.40 &\textbf{44.76} &\textbf{42.68} &\textbf{42.11} &\textbf{47.73}\\
	\rowcolor{black!10}\multirow{-2}*{+ COLA}& Last &87.79 &\textbf{62.79} &\textbf{45.94} &\textbf{44.17} &\textbf{41.90} &\textbf{41.31} &\textbf{47.22}\\
	\toprule[0.5pt]
	
	\multirow{2}*{NFGSM} &Best &80.43 &60.73 &51.09 &49.83 &48.92 &48.68 &51.85\\
	& Last &80.43 &60.73 &51.09 &49.83 &48.92 &48.68 &51.85\\
	\cline{2-9}
	\rowcolor{black!10}\multirow{2}*{+ COLA}&Best &80.06 &\textbf{61.90} &\textbf{52.46} &\textbf{51.70} &\textbf{50.84} &\textbf{50.63} &\textbf{53.50}\\
	\rowcolor{black!10}\multirow{-2}*{+ COLA}&Last &80.06 &\textbf{61.90} &\textbf{52.46} &\textbf{51.70} &\textbf{50.84} &\textbf{50.63} &\textbf{53.50}\\
	\toprule[0.5pt]
	
	\multirow{2}*{GAT} &Best &81.60 &64.34 &54.45 &53.87 &53.05 &52.74 &55.69\\
	& Last &82.44 &64.38 &54.33 &53.72 &52.73 &52.35 &55.50\\
	\cline{2-9}
	\rowcolor{black!10}\multirow{2}*{+ COLA}&Best &81.75 &\textbf{65.81} &\textbf{55.61} &\textbf{55.25} &\textbf{54.60} & \textbf{54.02} &\textbf{57.06}\\
	\rowcolor{black!10}\multirow{-2}*{+ COLA}&Last &81.99 &\textbf{65.89} &\textbf{55.61} &\textbf{55.13} &\textbf{54.40} &\textbf{53.93} &\textbf{56.99}\\
	\toprule[0.5pt]
	
	\multirow{2}*{FGSM-MEP} &Best &81.71 &65.02 &55.48 &55.26 &54.54 &54.38 &56.93\\
	& Last &81.71 &65.02 &55.48 &55.26 &54.54 &54.38 &56.93\\
	\cline{2-9}
	\rowcolor{black!10}\multirow{2}*{+ COLA}&Best &81.59 &\textbf{66.48} &\textbf{56.80} &\textbf{57.20} &\textbf{56.46} &\textbf{56.20} &\textbf{58.62}\\
	\rowcolor{black!10}\multirow{-2}*{+ COLA}&Last &82.19 &\textbf{66.56} &\textbf{56.43} &\textbf{56.69} &\textbf{55.69} &\textbf{55.38} &\textbf{58.15}\\
	\toprule[0.5pt]
	
	\multirow{2}*{TDAT} &Best &82.25 &66.15 &57.40 &56.85 &56.05 &55.90 &58.47\\
	&Last &82.25 &66.15 &57.40 &56.85 &56.05 &55.90 &58.47\\
	\cline{2-9}
	\rowcolor{black!10}\multirow{2}*{+ COLA}&Best &83.21 &\textbf{68.25} &\textbf{59.57} &\textbf{59.17} &\textbf{58.28} &\textbf{58.01} &\textbf{60.65}\\
	\rowcolor{black!10}\multirow{-2}*{+ COLA}&Last &83.20 &\textbf{67.82} &\textbf{58.57} &\textbf{58.98} &\textbf{57.79} &\textbf{57.50} &\textbf{60.13}\\
	\toprule[2pt]
\end{tabular}}
\label{pluginResults}
\end{table}

\subsection{Efficiency Analysis}\label{EffAnaSec}
\begin{figure}[t]\centering
	\subfigure{\includegraphics[scale=0.265]{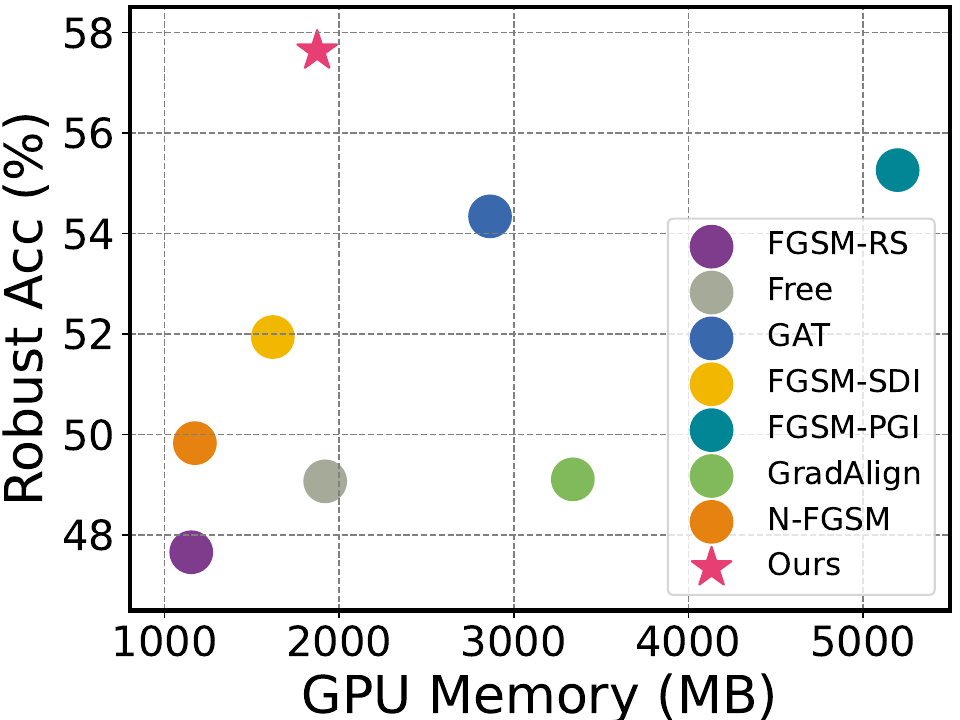}}
	\subfigure{\includegraphics[scale=0.265]{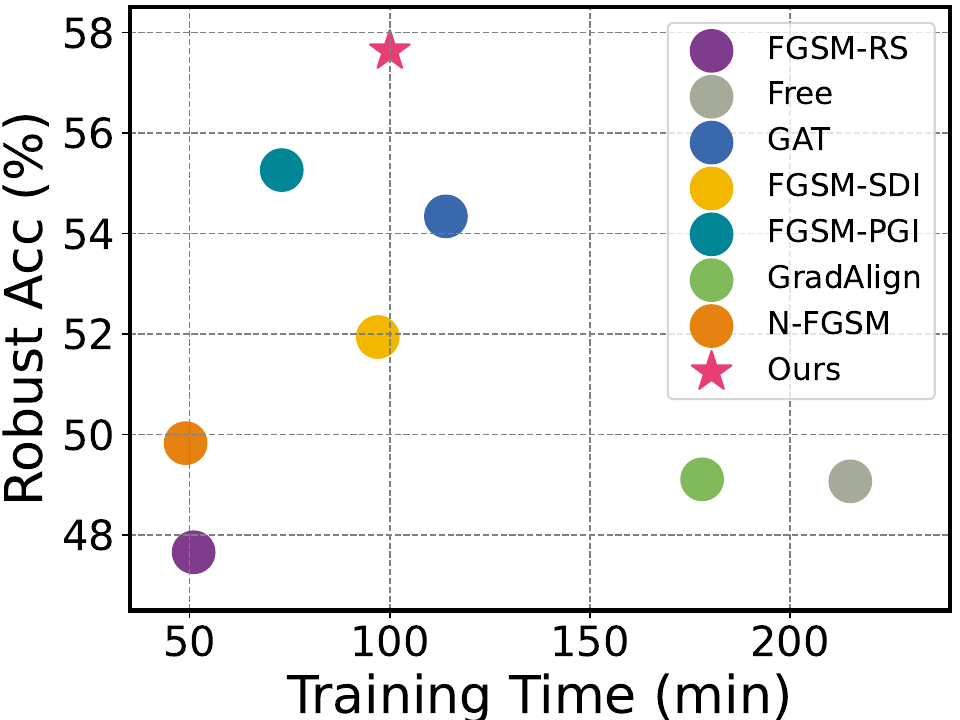}}
	\caption{Comparison of different methods regarding GPU memory usage, computation time, and PGD-10 robustness accuracy.}
	\label{EffAna}
\end{figure}
The computational complexity analysis is presented in Fig. \ref{EffAna}. We compared different methods in terms of GPU memory usage and computation time when training a ResNet-18 on CIFAR-10. The PGD-10 evaluates robust accuracy. The results show that our ETA achieves the highest robustness accuracy with relatively fewer training resources. Although there is a slight increase in memory usage, this significantly improves robustness accuracy. Memory usage increases due to the additional example count and storage. This issue can be mitigated by improving the training example loss discrimination scheme.

\subsection{Ablation Studies}\label{AblationSty}
\subsubsection{Effect of Each Component}
The contributions of each key component in ETA for robustness and clean accuracy are investigated in this subsection. The experiment is performed on the CIFAR-10 dataset and follows the default experimental settings. Table \ref{AblationResults} reports the corresponding ablation study results. Bold numbers indicate the best robust accuracy.
\begin{table}[b]
	\centering
	\caption{Ablation Study Results. BMI$^{\dagger}$, DLR$^{\ddagger}$, and TDL$^{\star}$ Represent Batch Momentum Initialization, Dynamic Label Relaxation, and Taxonomy Driven Loss.}
	\begin{tabular}[l]{@{}c c c c| c c}
	\toprule[2pt]
	\multirow{2}*{BMI$^{\dagger}$} & \multirow{2}*{DLR$^{\ddagger}$} & \multirow{2}*{TDL$^{\star}$} & \multirow{2}*{COLA} &Clean Acc & PGD-10\\
	&&&&Best/Last &Best/Last \\
	\toprule[1pt]
	\xmark &\xmark &\xmark &\xmark &64.29/91.33 &41.70/14.60\\
	\cmark &\xmark &\xmark &\xmark &82.50/82.91 &53.82/53.70\\
	\cmark &\cmark &\xmark &\xmark &86.73/89.04 &47.66/45.17\\
	\cmark &\xmark &\cmark &\xmark &83.83/84.10 &51.00/50.55\\
	\cmark &\cmark &\cmark &\xmark &82.25/82.25 &56.85/56.85\\
	\cmark &\cmark &\cmark &\cmark &83.19/83.21 &\textbf{57.63/57.31}\\
	\toprule[2pt]
\end{tabular}
\label{AblationResults}
\end{table}
\begin{figure}[t]\centering
	\subfigure[CIFAR-100]{\includegraphics[scale=0.265]{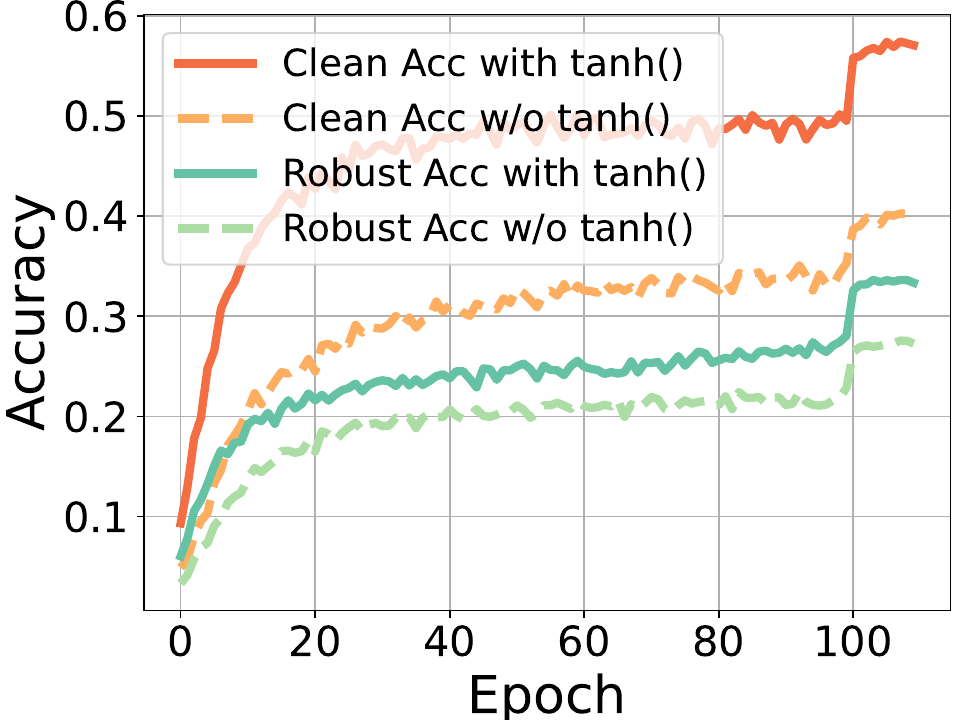}}
	\subfigure[ImageNet-100]{\includegraphics[scale=0.265]{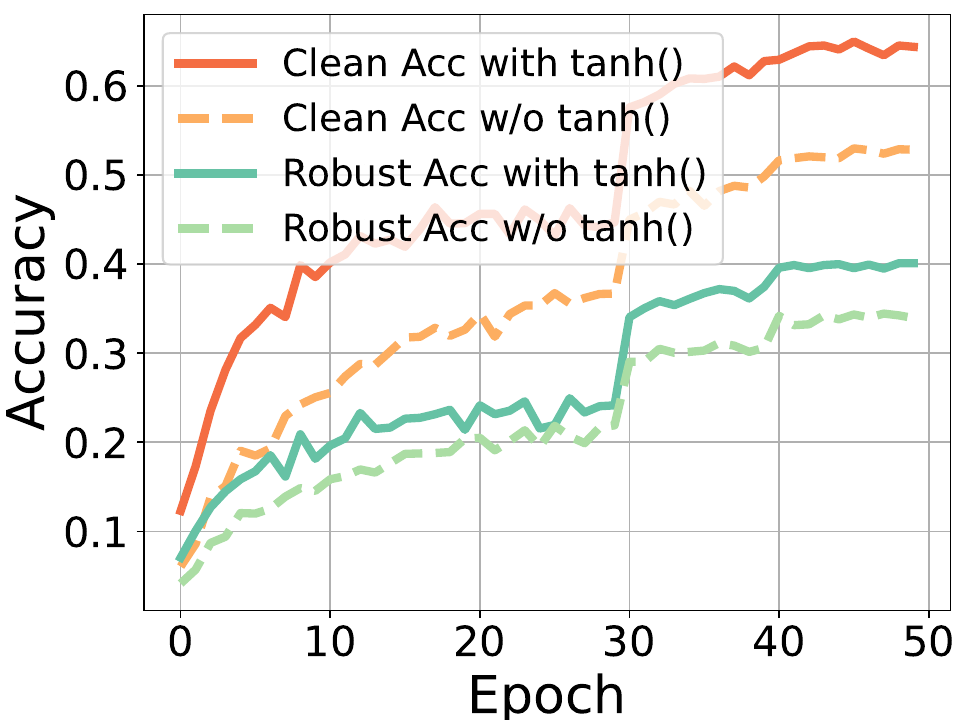}}
	\caption{Performance comparison in cases of with or w/o $\text{tanh}(\cdot)$ in taxonomy driven loss \eqref{TDLDef}.}
	\label{tanhfig}
\end{figure}
First, batch momentum initialization enhances training stability and eliminates catastrophic overfitting. This is achieved by increasing the diversity of the distribution of AEs. Nonetheless, the robust accuracy is unsatisfactory and does not achieve state-of-the-art performance. Second, the clean accuracy is improved by using batch momentum initialization and dynamic label relaxation. However, this combination compromises the robust accuracy of the model. Third, using batch momentum initialization and taxonomy-driven loss without dynamic label relaxation fails to mitigate the negative impact of AEs generated from misclassified examples, resulting in a degradation of robust accuracy. However, when batch momentum initialization, dynamic label relaxation, and taxonomy-driven loss are used together, clean and robust accuracy achieves better trade-offs and is significantly enhanced. Finally, adding the COLA further enhances clean and robustness accuracy, resulting in the best training performance.
\subsubsection{Effect of $\text{tanh}(\cdot)$ in Taxonomy Driven Loss}
To analyze the effect of $\text{tanh}(\cdot)$ in taxonomy driven loss, the corresponding results on CIFAR-100 and ImageNet-100 datasets are shown in Fig. \ref{tanhfig}. In detail, by introducing $\text{tanh}(\cdot)$ for acceleration, the taxonomy driven loss improves the clean accuracy of the model in the early stages of training. In contrast, the training has limited improvement in clean accuracy during the early epochs, in the case of those without tanh, which subsequently affects the final training performance. For the CIFAR-100 dataset, the difference in clean accuracy with and w/o $\text{tanh}(\cdot)$ reaches $17\%$, while the robust accuracy reaches $6.08\%$. For the ImageNet-100 dataset, the gaps in clean and robust accuracy at the final checkpoint are $11.6\%$ and $6.2\%$, respectively. This supports the motivation behind the dynamic label relaxation, which is presented in subsection \ref{DLRDefSec}.

\subsection{Effects of with Different Epochs and Perturbation Budgets}
\begin{figure}[b]\centering
	\subfigure{\includegraphics[scale=0.26]{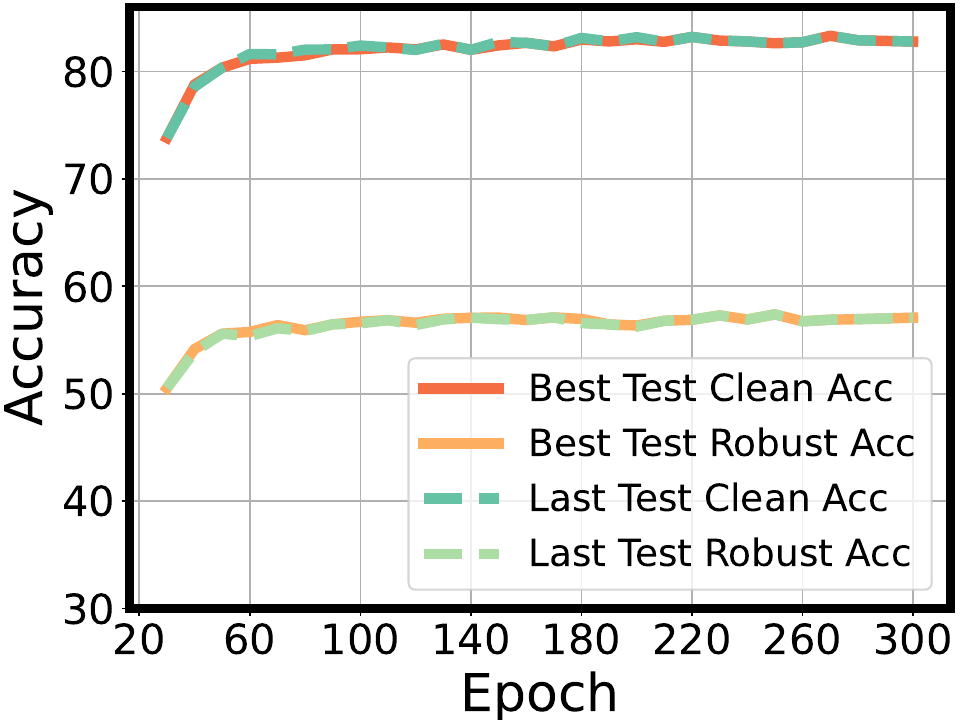}}
	\subfigure{\includegraphics[scale=0.26]{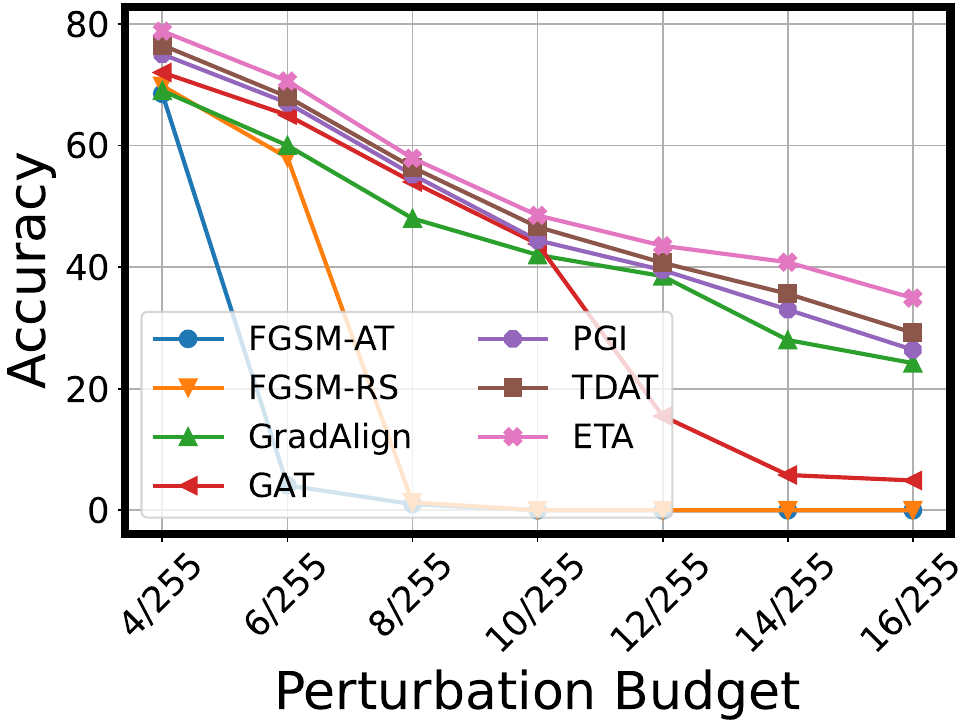}}
	\caption{Comparison of different training epochs and effect of perturbation budget. (Left) Performance with different training epochs. (Right) Performance of different methods under PGD-10 with different perturbation budgets.}
	\label{PDEPDF}
\end{figure}
\subsubsection{Different Epochs}
This part presents the performance of our ETA using different epochs to investigate model training stability, which is an important indicator of methods \cite{ZZGGES, SPHXL, WJSAD}. The robust accuracy is evaluated by PGD-10. We compare training effects from 30 to 300 training epochs at intervals of 10 epochs, and the results are shown in Fig. \ref{PDEPDF} (Left). Specifically, the last and best accuracies of our ETA only possess minor gaps under different training epoch settings, with the best model performance almost coinciding with the performance at the end of the training. This observation demonstrates the stability of our method, indicating that it can avoid comparing different checkpoints. These results also verify the ability of our method to address catastrophic overfitting, as the training stability does not collapse even with more training epochs. Furthermore, the performance of our ETA can be further enhanced as the training epochs increase.

\subsubsection{Different Perturbation Budgets}
We follow the experimental settings of \cite{PGIMEP} and consider other FAT approaches on the CIFAR-10 dataset with ResNet18 as the backbone to analyze the performance of our method under different perturbation budgets for training and evaluation. The robust accuracy is evaluated by PGD-10, and the results are presented in Fig. \ref{PDEPDF} (Right). Previous FAT methods suffer when training with larger perturbation budgets (as in FGSM-AT, FGSM-RS, and GAT). Conversely, our proposed ETA can eliminate catastrophic overfitting and significantly improve robust accuracy.

\begin{figure}[b]\centering
	\subfigure[]{\includegraphics[scale=0.265]{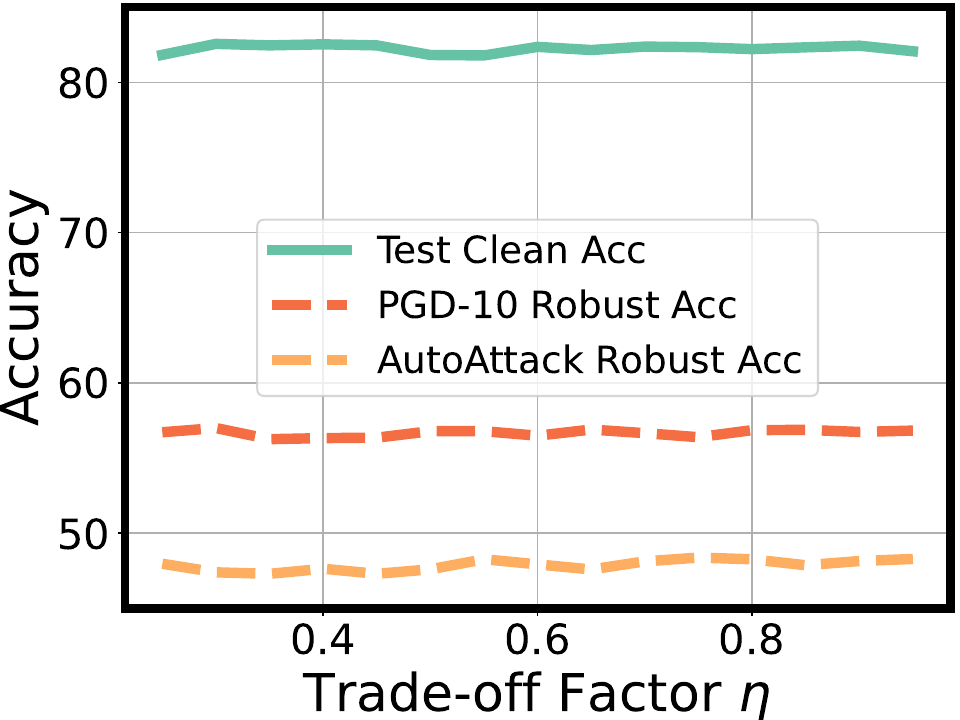}}
	\subfigure[]{\includegraphics[scale=0.265]{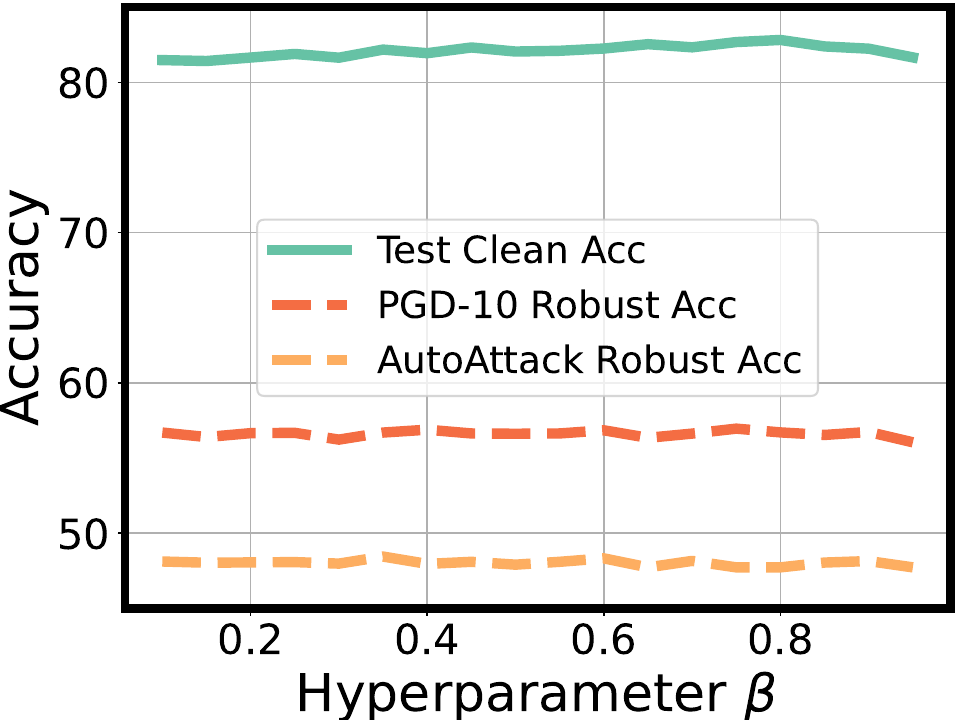}}
	\subfigure[]{\includegraphics[scale=0.265]{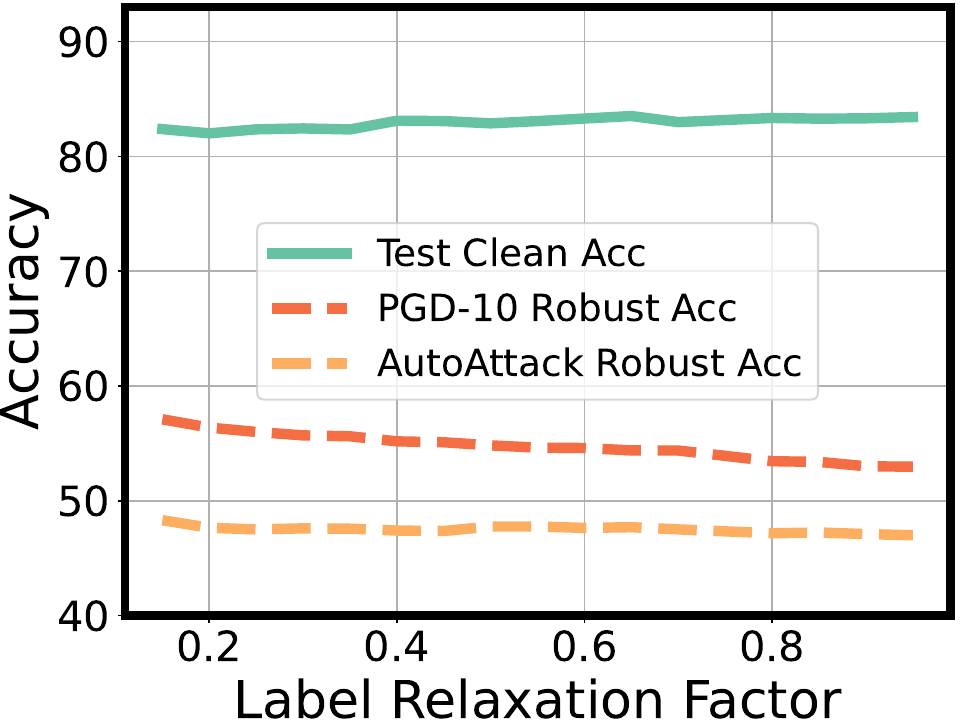}}
	\subfigure[]{\includegraphics[scale=0.265]{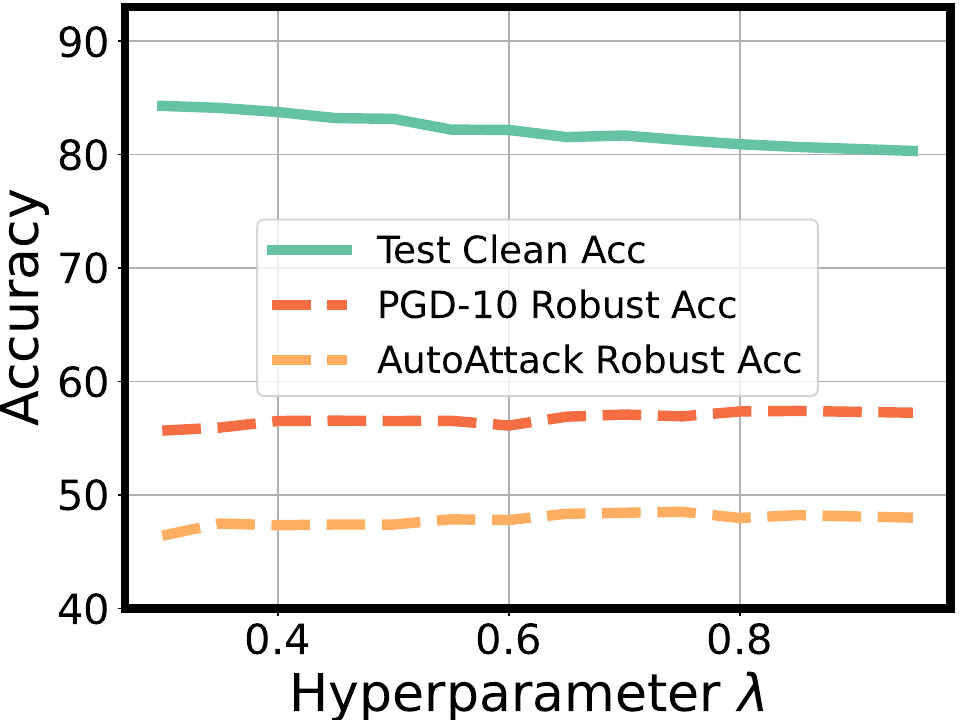}}
	\caption{Hyperparameters selection results. (a) Perturbation momentum trade-off factor in \eqref{BMIOne}--\eqref{BMITwo}. (b) Hyperparameter $\beta$ in \eqref{DLRDetail}. (c) Minimum relaxation factor $\gamma_\text{min}$ in \eqref{DLRDetail}. (d) Regularization hyperparameter $\lambda$ in \eqref{TDLDef}. }
	\label{HyperFigs}
\end{figure}

\begin{table}[t]
	\centering
	\caption{COLA with Compress Losses}
	\setlength{\tabcolsep}{1.3mm}{
	\begin{tabular}[l]{c| c| c| c c c c c c}
	\toprule[2pt]
	Factor & &Clean &FGSM &MI &PGD10 &PGD20 &PGD50 &RA\\
	\toprule[1pt]
	\multirow{2}*{0.9} &Best &82.32 &66.99 &57.54 &57.05 &56.47 &56.28 &58.86\\
	& Last &82.38 &66.65 &57.36 &56.90 &56.18 &55.96 &58.61\\
	\toprule[0.5pt]
	\multirow{2}*{0.8} &Best &83.19 &66.98 &57.89 &57.63 &56.83 &56.63 &59.19\\
	& Last &83.21 &66.69 &57.66 &57.31 &56.41 &56.22 &58.85\\
	\toprule[0.5pt]
	\multirow{2}*{0.7} &Best &82.51 &67.20 &58.59 &58.11 &57.34 &57.06 &59.66\\
	& Last &82.71 &66.79 &58.01 &57.63 &56.95 &56.88 &59.25\\
	\toprule[0.5pt]
	\multirow{2}*{0.6} &Best &82.85 &67.12 &58.34 &58.00 &57.27 &57.11 &59.56\\
	& Last &83.12 &67.34 &56.95 &57.74 &57.20 &57.00 &59.24\\
	\toprule[0.5pt]
	\multirow{2}*{0.5} &Best &83.11 &68.25 &59.57 &59.17 &58.28 &58.01 &60.65\\
	& Last &83.05 &67.54 &58.82 &58.44 &57.62 &57.34 &59.95\\
	\toprule[0.5pt]
	\multirow{2}*{0.4} &Best &82.51 &67.93 &59.47 &59.30 &58.51 &58.34 &60.71\\
	& Last &82.55 &67.67 &59.18 &58.83 &57.98 &57.78 &60.28\\
	\toprule[0.5pt]
	\multirow{2}*{0.3} &Best &82.16 &68.24 &60.36 &60.04 &59.27 &58.98 &61.37\\
	& Last &82.25 &67.83 &59.76 &59.42 &58.61 &58.29 &60.78\\
	\toprule[0.5pt]
	\multirow{2}*{0.2} &Best &81.50 &68.70 &61.89 &61.35 &60.64 &60.25 &62.56\\
	& Last &81.69 &68.63 &61.33 &61.01 &60.15 &60.00 &62.22\\
	\toprule[0.5pt]
	\multirow{2}*{0.1} &Best &80.14 &69.00 &63.42 &62.53 &61.43 &61.21 &63.51\\
	& Last &80.06 &68.97 &63.31 &62.25 &61.41 &61.14 &63.41\\
	\toprule[2pt]
\end{tabular}}
\label{DecFactors}
\end{table}

\begin{figure*}[t]\centering
	\subfigure[FGSM-RS]{\includegraphics[scale=0.109]{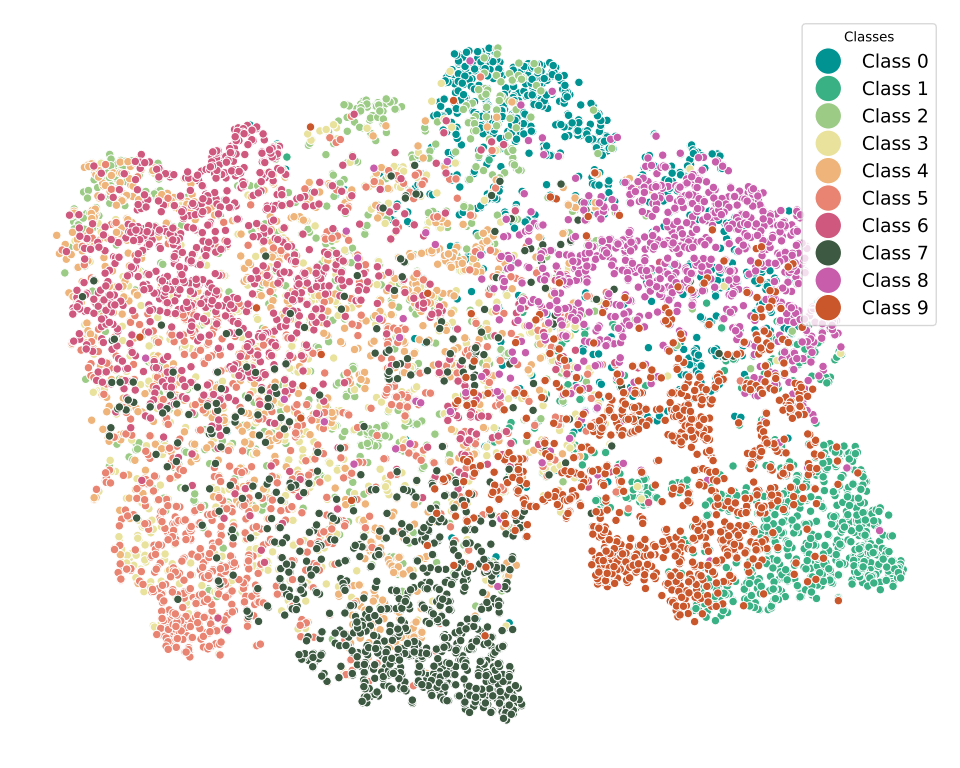}}
	\subfigure[GAT]{\includegraphics[scale=0.109]{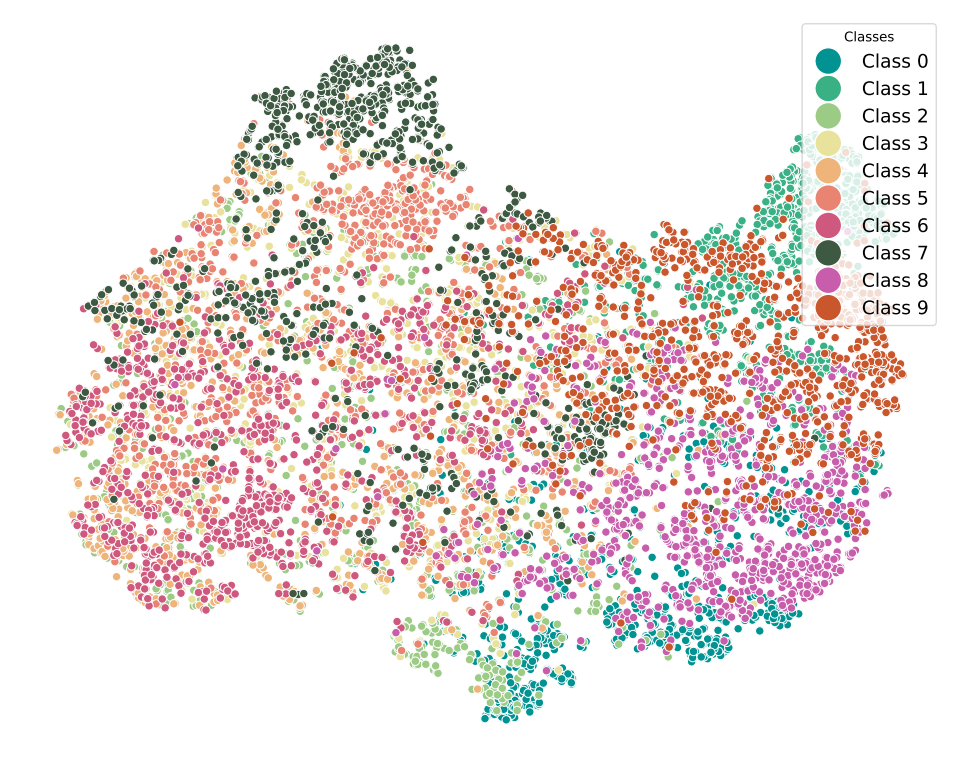}}
	\subfigure[NuAT]{\includegraphics[scale=0.109]{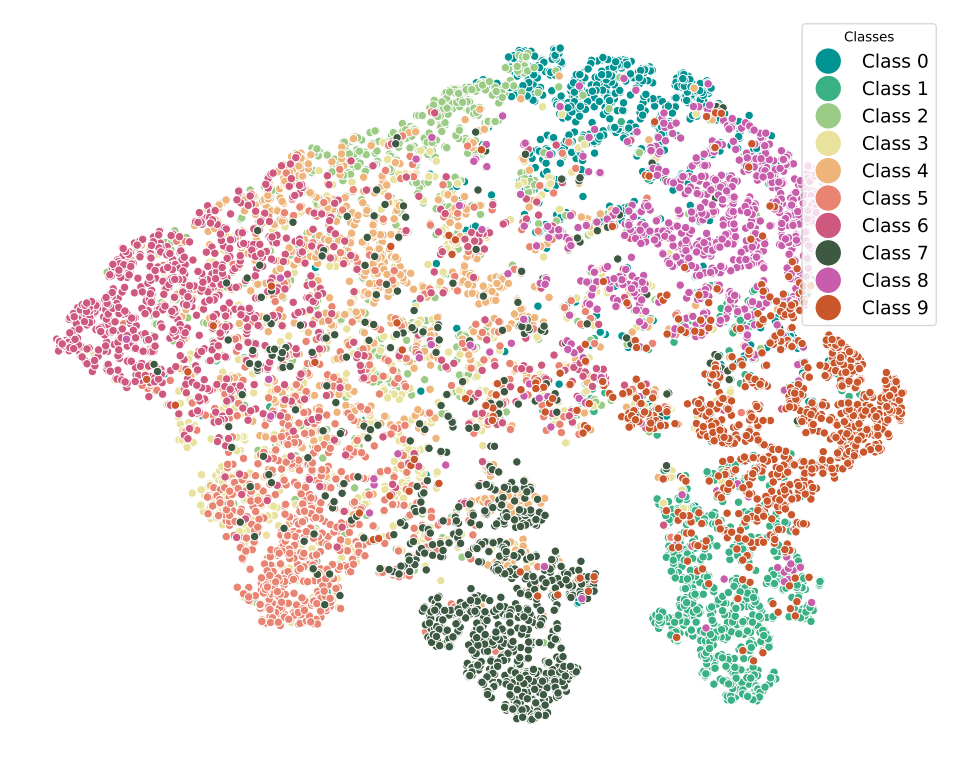}}
	\subfigure[GradAlign]{\includegraphics[scale=0.109]{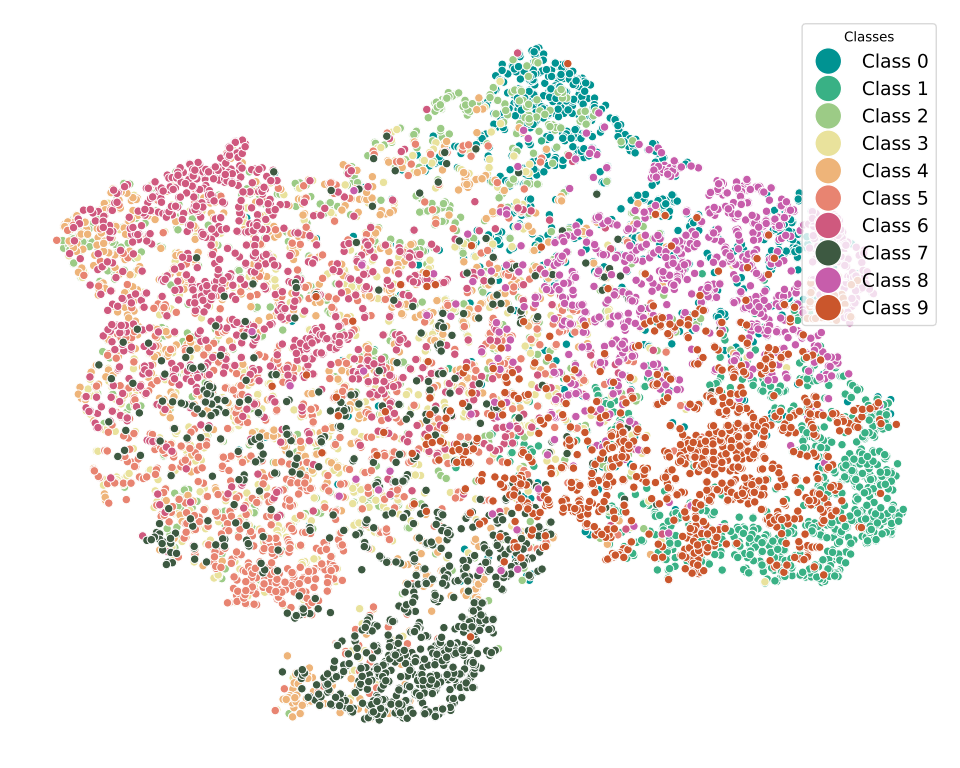}}
	\subfigure[NFGSM]{\includegraphics[scale=0.109]{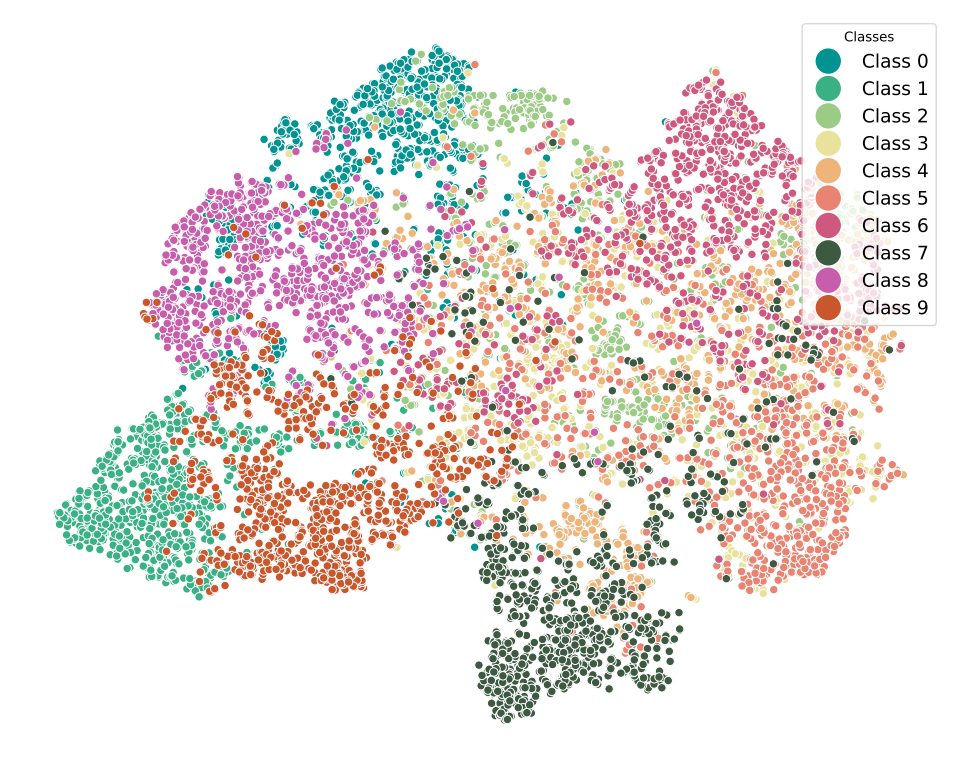}}
	\subfigure[Ours]{\includegraphics[scale=0.109]{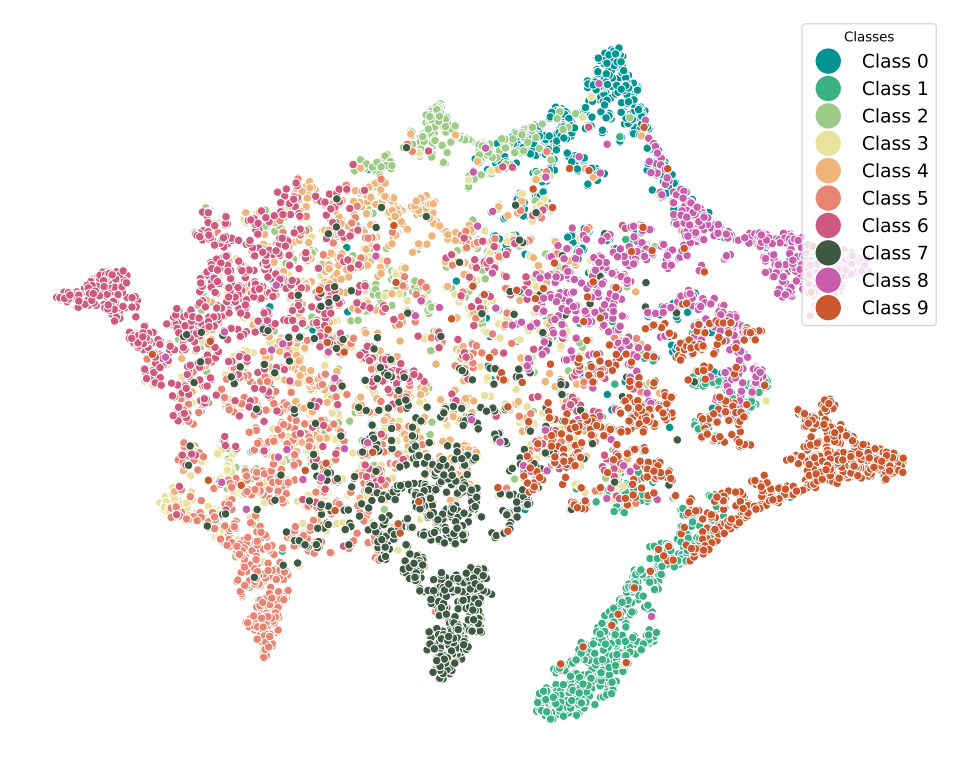}}
	\caption{Visualizations of t-SNE for different methods classification performance of AEs on CIFAR-10.}
	\label{TSNEAE}
\end{figure*}

\begin{figure}[t]\centering
	\subfigure[FGSM-RS]{\includegraphics[scale=0.15]{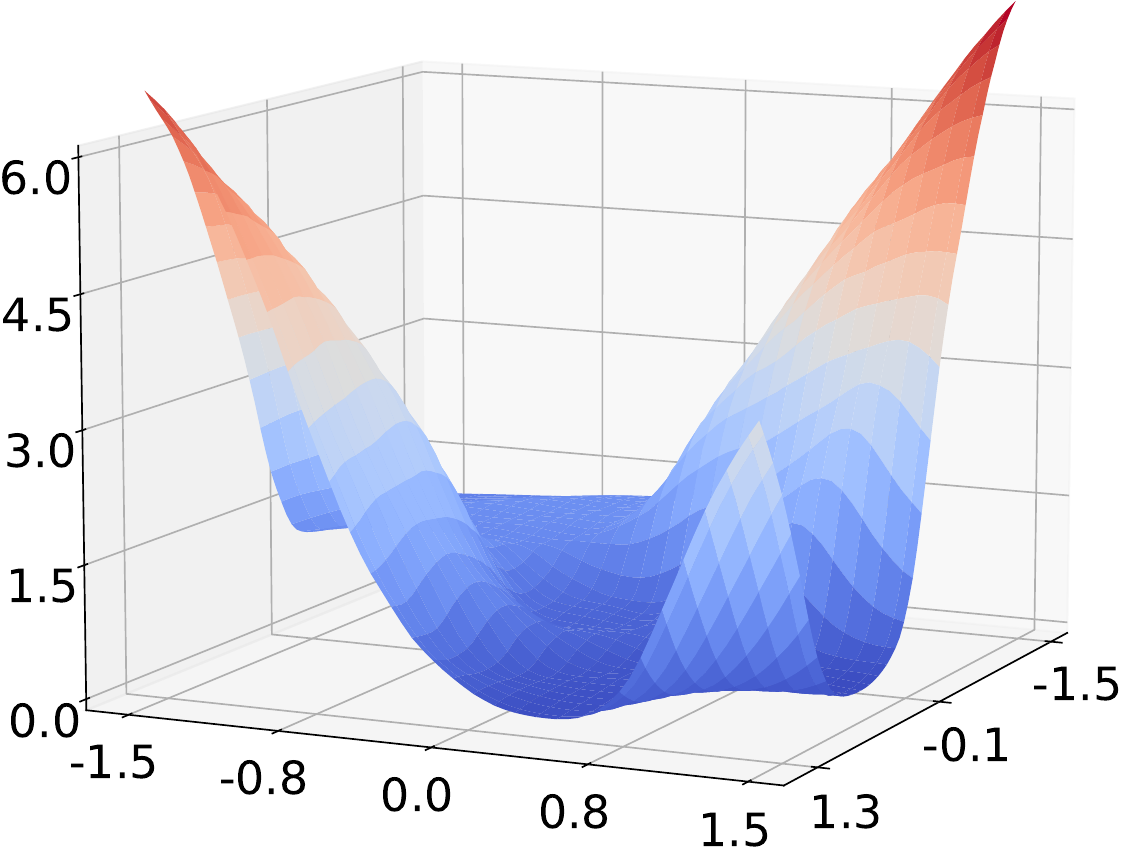}}
	\subfigure[GradAlign]{\includegraphics[scale=0.15]{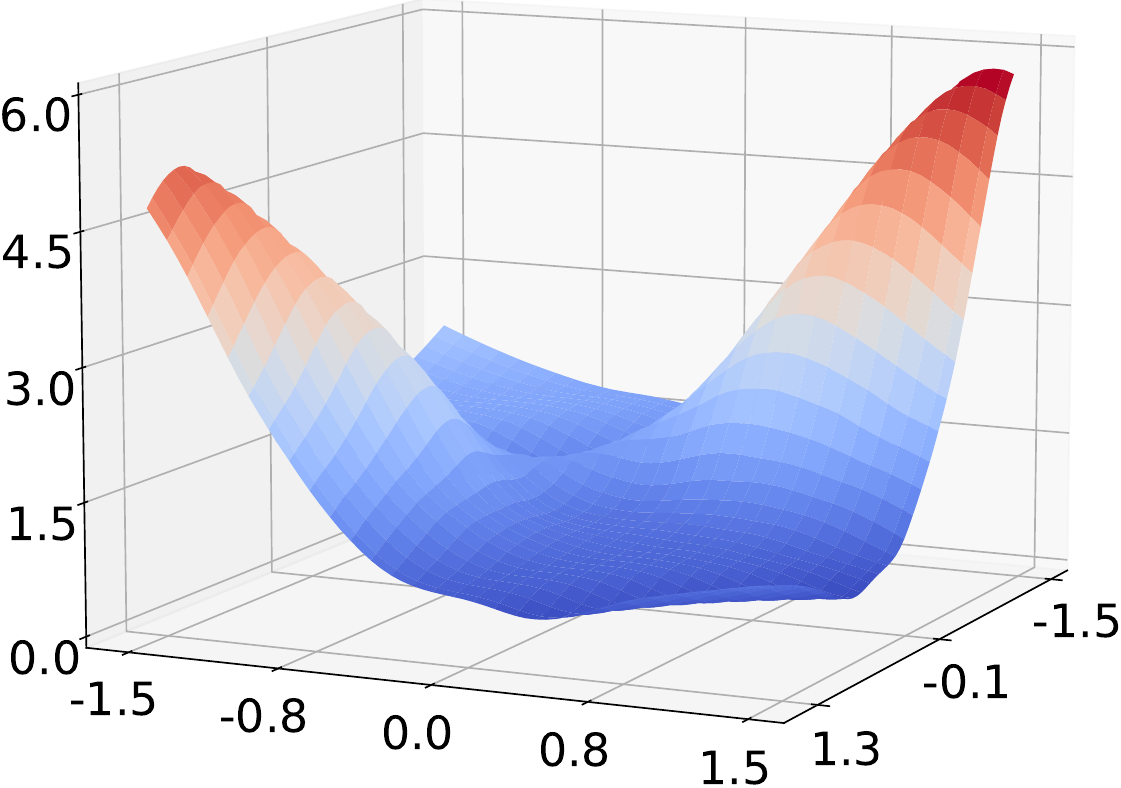}}
	\subfigure[GAT]{\includegraphics[scale=0.15]{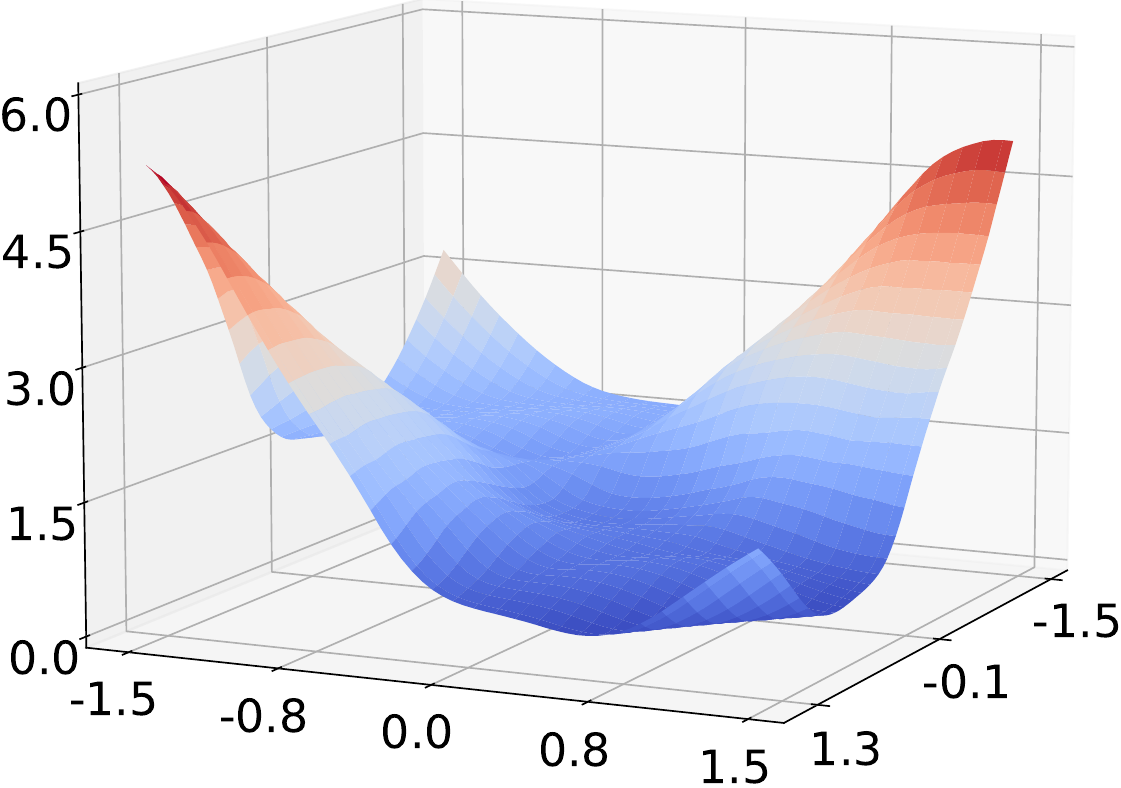}}
	\subfigure[Free-AT]{\includegraphics[scale=0.15]{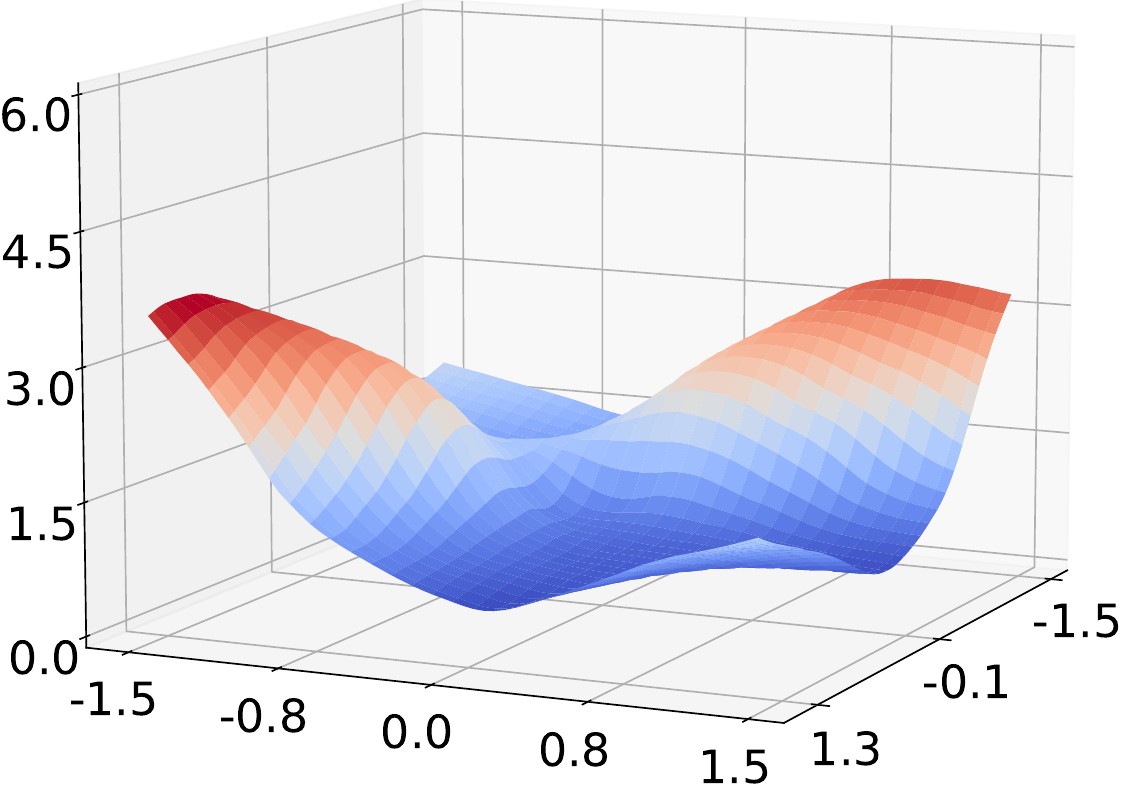}}
	\subfigure[FGSM-SDI]{\includegraphics[scale=0.15]{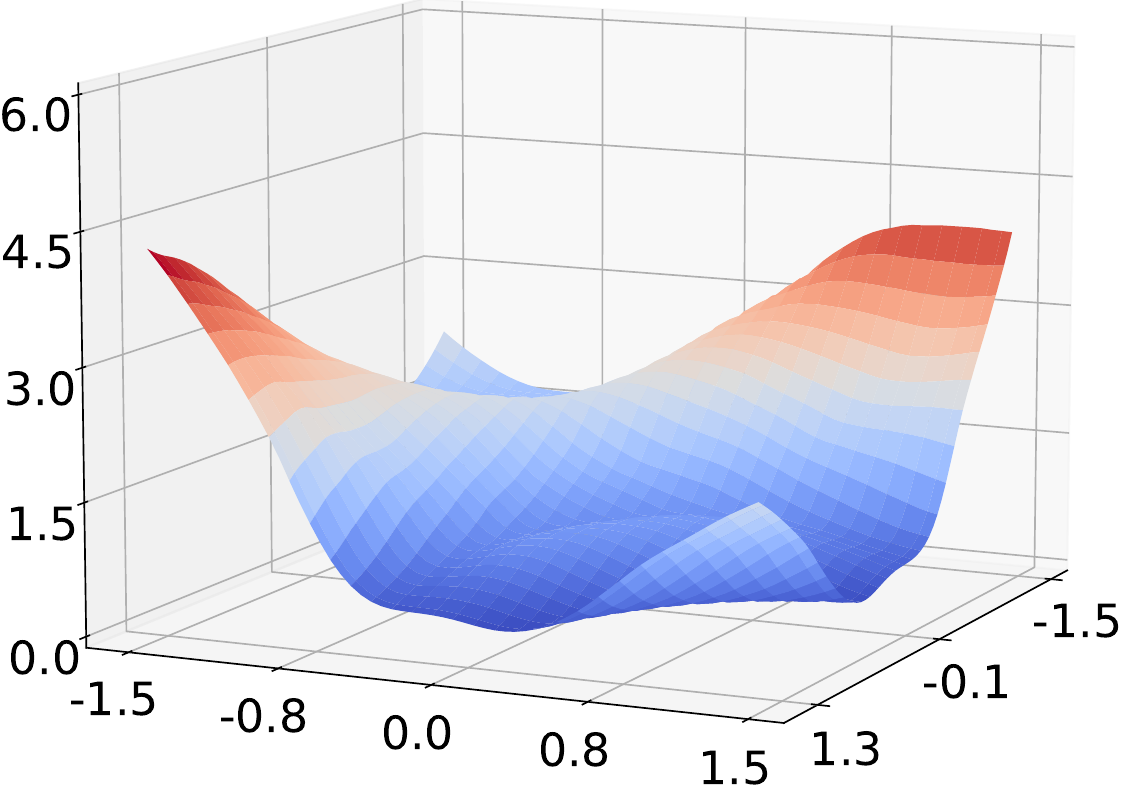}}
	\subfigure[NFGSM]{\includegraphics[scale=0.15]{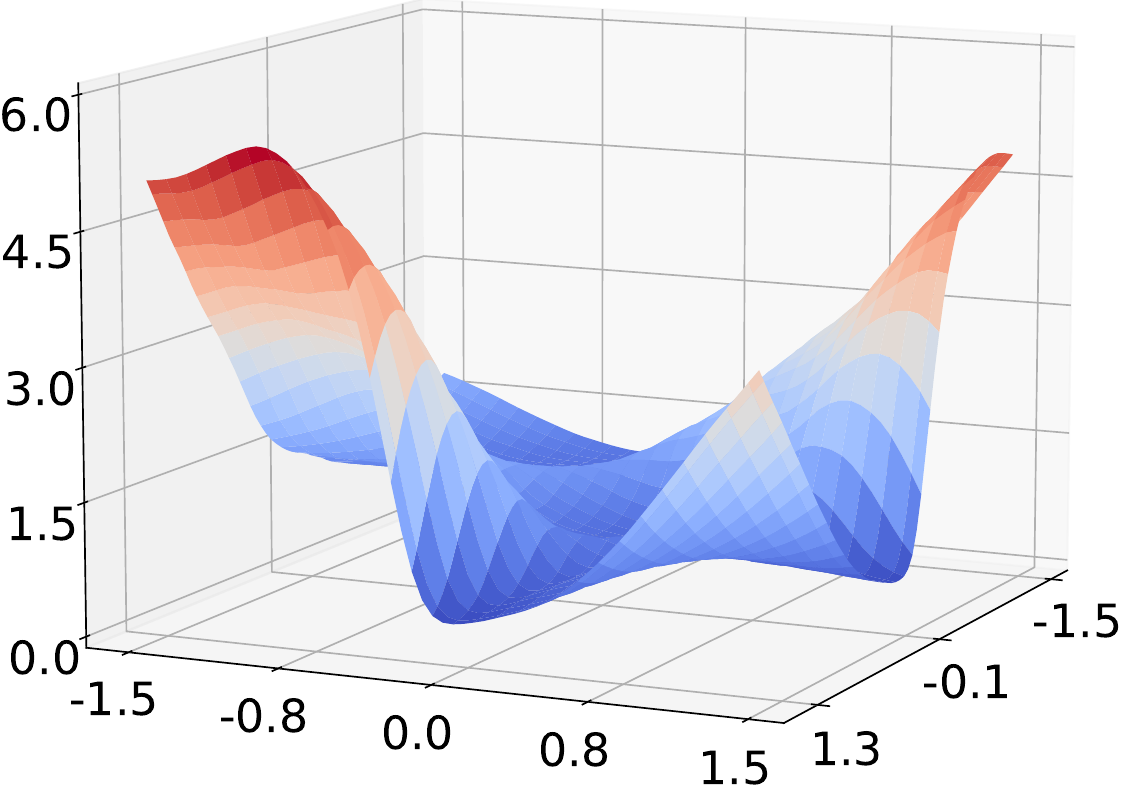}}
	\subfigure[MART]{\includegraphics[scale=0.15]{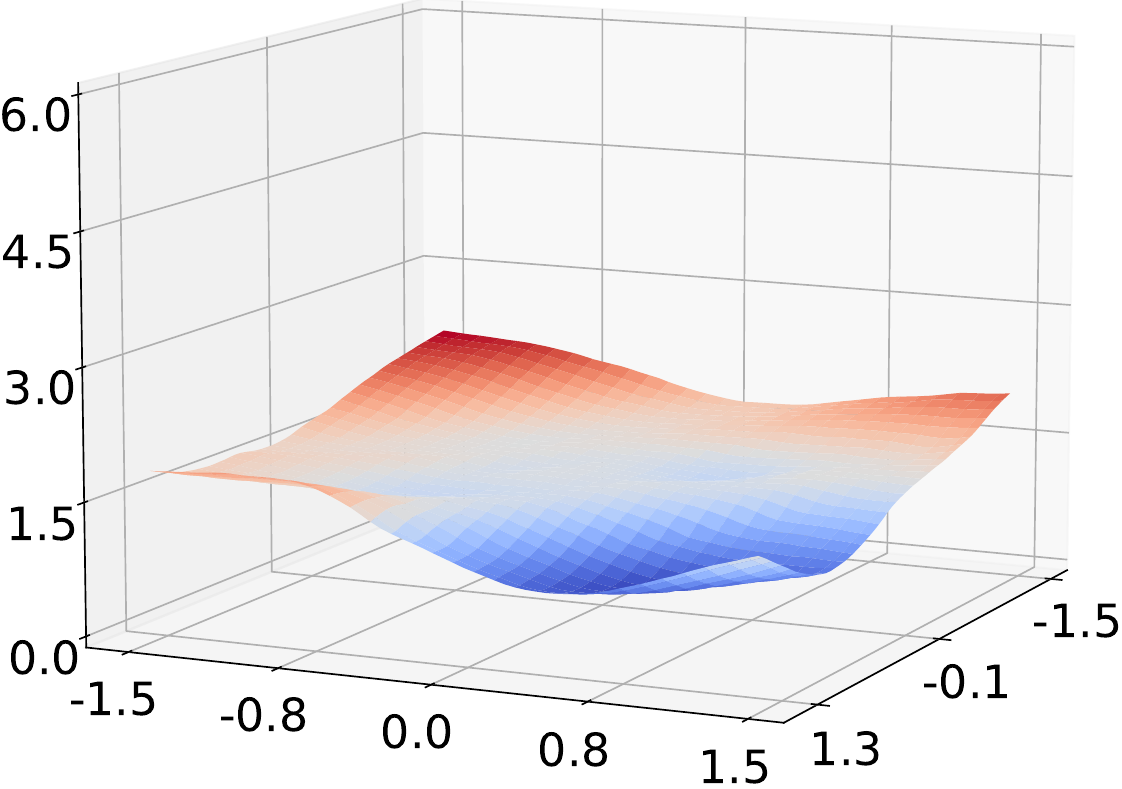}}
	\subfigure[FGSM-MEP]{\includegraphics[scale=0.15]{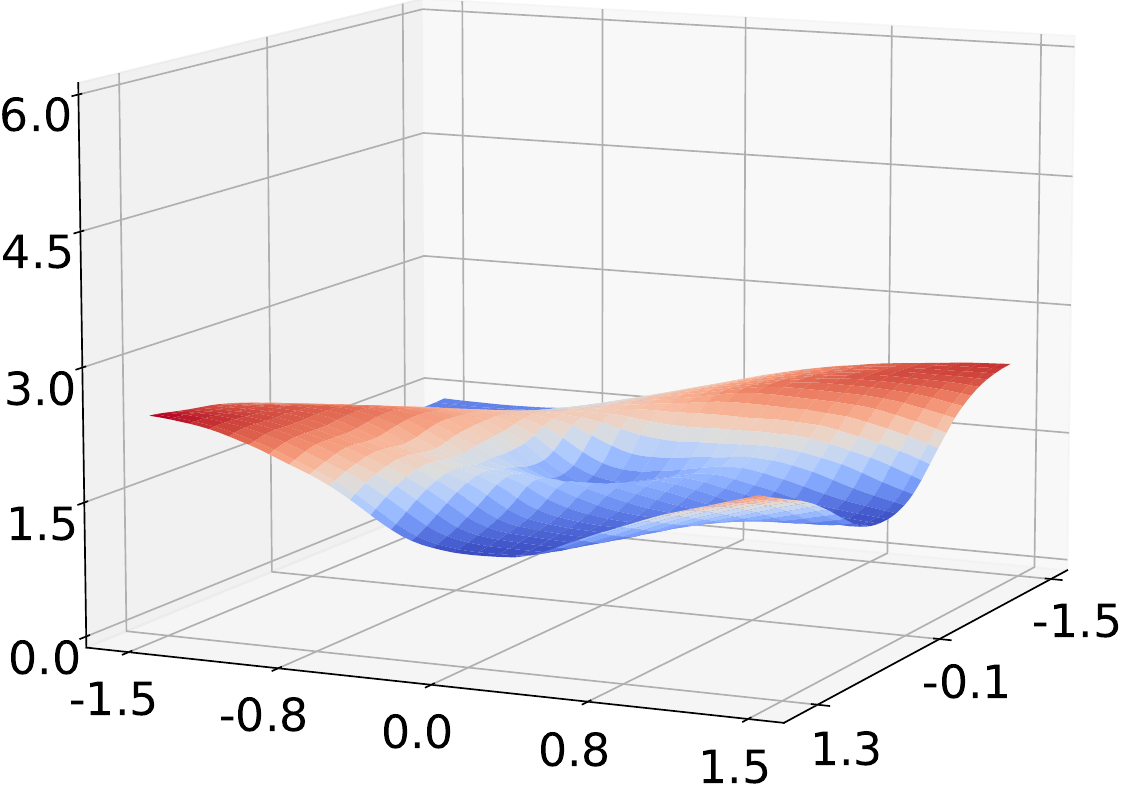}}
	\subfigure[Ours]{\includegraphics[scale=0.15]{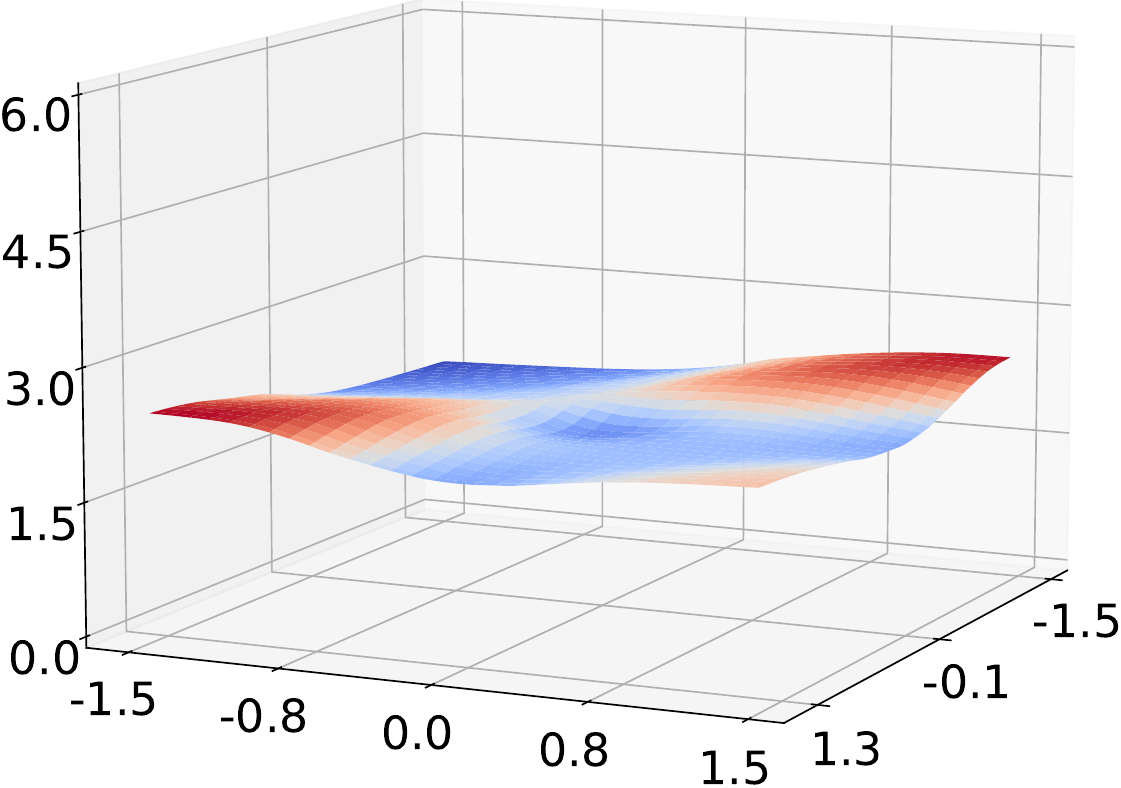}}
	\caption{Comparison of different methods loss surface on CIFAR10.}
	\label{LossSurResults}
\end{figure}

\subsection{Hyper-parameters Selection}
We perform hyper-parameter influence experiments to analyze the importance of hyper-parameters in our proposed method to robustness. Our proposed ETA involves four key hyper-parameters, including perturbation trade-off factor $\eta$ in \eqref{BMIOne} - \eqref{BMITwo}, relaxation amplitude factor $\beta$ in \eqref{DLRDetail}, regularization trade-off factor in \eqref{TDLDef}, and loss adaptation factor $\eta_c$ in \eqref{LCGDef}. The control variable approach is adopted to identify the appropriate hyper-parameters. We adopt the PGD-10 and AutoAttack to evaluate the robust accuracy
\subsubsection{Effect of Perturbation Trade-off Factor}
To evaluate the influence of the perturbation trade-off factor, we execute experiments adopting ETA with varying value factors. The results are presented in Fig. \ref{HyperFigs} (a). Specifically, this component is not sensitive to hyperparameters, and the robust and clean accuracy is inversely and directly proportional to the parameter $\alpha$, respectively. This component obtains the best robust accuracy with a trade-off factor set to $0.75$.

\subsubsection{Effect of Label Relaxation Factor}
Fig. \ref{HyperFigs} (b) presents the results of analyzing the effect of the relaxation factor $\beta$ in dynamic label relaxation \eqref{DLRDef}. We sample 35 factors ranging from $0.11$ to $1$ to evaluate the robust accuracy. In detail, robust accuracy gradually decreases as the relaxation factor increases, while clean accuracy gradually increases. Nevertheless, our ETA maintains a clean accuracy of over $82.4\%$ on the CIFAR-10 when achieving the best robust accuracy. Therefore, setting a smaller factor can achieve the trade-off between clean and robust accuracy. Note that using different factors does not lead to catastrophic overfitting, emphasizing the stability of our method to training. Besides, we provide a series of experiments to assess the impact of minimum relaxation bound $\gamma_\text{min}$ with the results presented in Fig. \ref{HyperFigs}(c). Our ETA achieves the best robustness with the value of $\gamma_\text{min}$ set to 0.15.

\subsubsection{Effect of Hyperparameter in Taxonomy Driven Loss}
To identify the influence of regularization hyperparameter $\lambda$, we execute a series of experiments using the proposed ETA with varying hyperparameter values. The experiment results involving the regularization hyperparameter $\lambda$ are shown in Fig. \ref{HyperFigs}(d). Our method achieves the best adversarial robustness with $\lambda=0.75$ across all adversarial attack situations.

\subsubsection{Effects of Factor in COLA}\label{EFCOLA}
This part analyzes the impact of the scaling factor in COLA with the results shown in Table \ref{DecFactors}, illustrating the impact of different degrees of example loss compression on training. This table shows that as the loss adjustment factor decreases, the ability to resist gradient-based adversarial attacks gradually improves. Specifically, robustness accuracy significantly increases under FGSM (+2.85$\%$), MIFGSM (+6.02$\%$), PGD-10 (+5.68$\%$), PGD-20 (+5.38$\%$), and PGD-50 (+5.31$\%$). Simultaneously, clean accuracy shows an upward trend as the loss adjustment factor decreases from 0.8 to 0.5. However, clean accuracy declines with a further reduction of the loss adjustment factor.

\vspace{-0.3cm}
\subsection{Visualization Analysis}\label{VAResults}
\subsubsection{Visualization of t-SNE}
To visually demonstrate the comparison of our and other methods, we employed the t-SNE technique for visualization in this section. The results are shown in Fig. \ref{TSNEAE}. The visualization experiments adopt the best checkpoint obtained for each method. Outputs of model on the CIFAR-10 test set are used as vectors, resulting in a matrix of size $10000\times10$. These vectors are reduced in dimensionality using the principal component analysis and then visualized using the t-SNE. Our ETA achieves the best classification performance on AEs. Specifically, the distances between the prediction results of the model for different examples within the same class are closer, while the outputs for different classes are sufficiently distant from each other. This indicates that the regions corresponding to our method are more compact and have higher discriminative power. These findings highlight the competitiveness of our approach in effectively distinguishing between AEs, thereby improving the robustness.
\subsubsection{Loss Surface}
A flatter loss landscape indicates that the model is more robust against AEs \cite{AWPHRG, PGK}. Therefore, we investigate the adversarial loss landscapes of our ETA and other methods. The results are demonstrated in Fig. \ref{LossSurResults}. We change the model input along a linear space defined by a random Rademacher direction and the adversarial direction discovered by PGD-10 to generate loss landscapes. The figure shows that our method exhibits a more linear cross-entropy loss along the adversarial direction than other FAT methods. This verifies that the proposed ETA better preserves the local linearity of the model. These results highlight the competitiveness of the ETA in obtaining a flatter adversarial loss landscape.

\subsubsection{Example Loss Value}
\begin{figure}[t]\centering
	\subfigure{\includegraphics[scale=0.265]{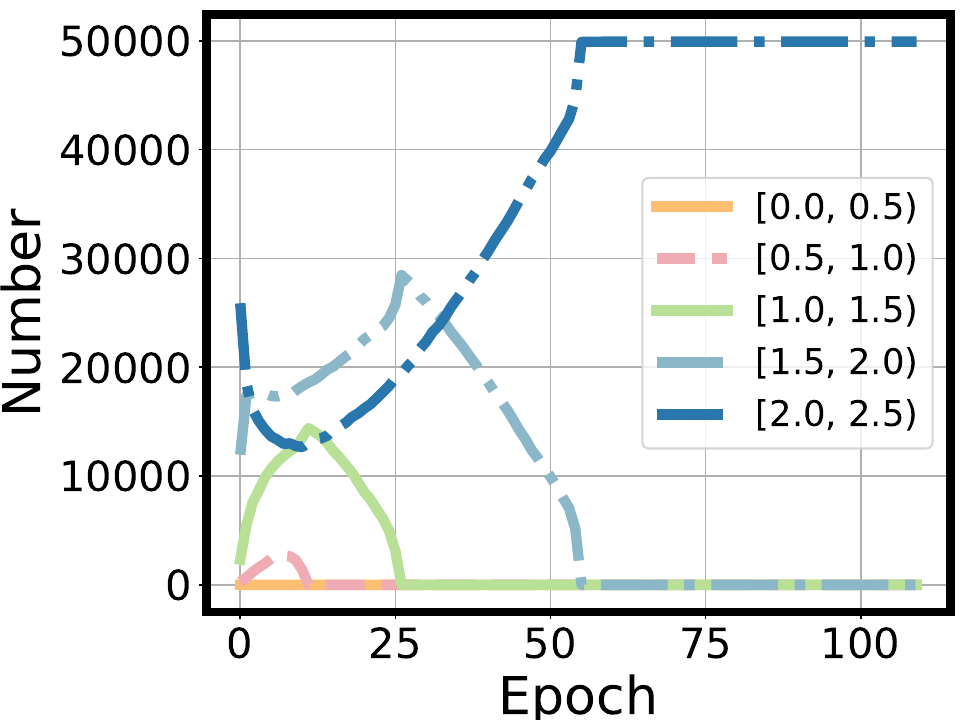}}
	\subfigure{\includegraphics[scale=0.265]{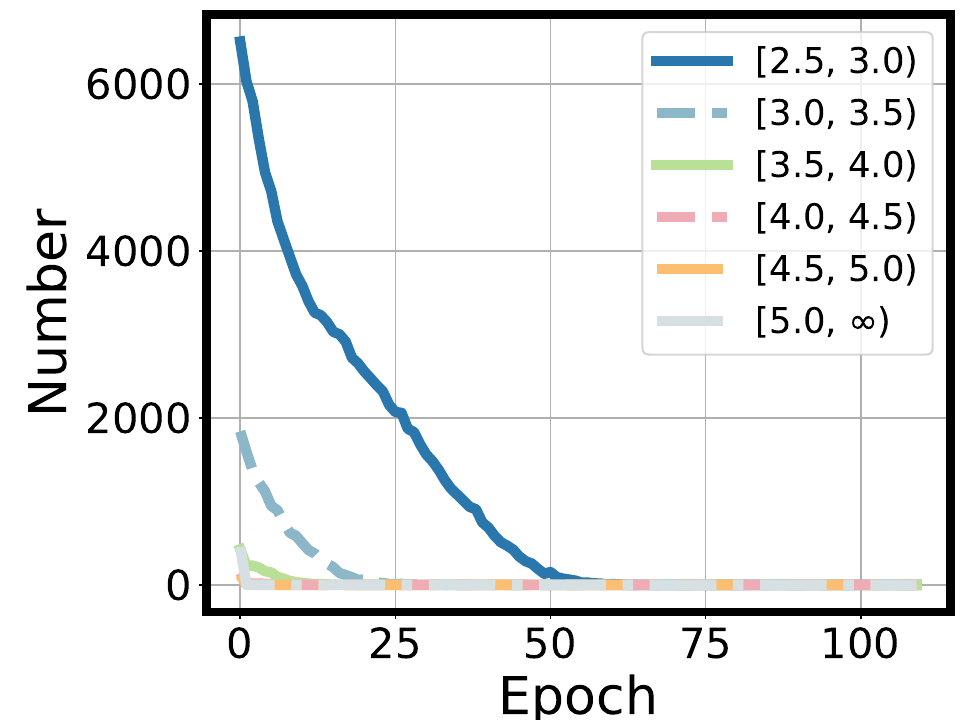}}
	\caption{The change in the number of examples in different loss intervals when performing our ETA with ResNet18.}
	\label{LCTAEE}
\end{figure}
To analyze the reason why our ETA improves the distribution of example exploitation and thereby enhances the performance of adversarial training, we present the changes in the number of examples with different losses in Fig. \ref{LCTAEE} during the execution of ETA on the CIFAR-10 with perturbation budget $\epsilon=8/255$. The results show that the training example losses are more concentrated, and as training progresses, almost all example losses are concentrated between 2 and 2.5. This conclusion highlights that our method stabilizes the training process. The balanced inner and outer optimization process of the min-max paradigm in training is why ETA can eliminate catastrophic overfitting and enhance robustness.

\section{Conclusion and Outlook}\label{ConOut}
In this paper, we present a taxonomy to explain and understand catastrophic overfitting. Results reveal an issue of label flipping where the effectiveness of adversarial examples significantly degenerates, which leads to catastrophic overfitting. Based on the taxonomy, a strong connection between label flipping and adversarial examples generated from misclassified clean examples is found. To further analyze the catastrophic overfitting during optimization, we propose to investigate the magnitude of training losses. The results show extreme training loss can lead to catastrophic overfitting and negatively impact the robustness of the model, which motivates us to redesign the loss function. We introduce batch momentum initialization to improve the diversity of AES for relieving the min-max imbalance in fast adversarial training (FAT). To mitigate the negative impact of AEs generated from misclassified clean examples, we develop dynamic label relaxation and taxonomy-driven loss. Additionally, we present catastrophic overfitting aware loss adaptation (COLA) to centralize the losses, further enhancing training performance and eliminating catastrophic overfitting. Extensive evaluations conducted on four benchmark databases demonstrate that adopting our ETA can alleviate catastrophic overfitting and achieve better robust accuracy than existing methods. Our proposed COLA can also be used as a plug-in to improve the performance of existing FAT. Our future works will focus on the following aspects. First, the strength of the attack restricts the performance of the FAT. Therefore, leveraging our method to perform adversarial training with a more powerful attack could be valuable. Second, the impact of example category imbalance can be analyzed from the perspective of refined taxonomy, and further be utilized to develop improved adversarial training.

\bibliographystyle{IEEEtran}
\bibliography{Manuscript}{}


\begin{IEEEbiography}[{\includegraphics[width=1in,height=1.25in,clip,keepaspectratio]{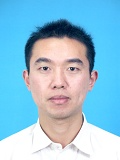}}]{Jie Gui} (SM'16) is currently a professor at the School of Cyber Science and Engineering, Southeast University. He received a BS degree in Computer Science from Hohai University, Nanjing, China, in 2004, an MS degree in Computer Applied Technology from the Hefei Institutes of Physical Science, Chinese Academy of Sciences, Hefei, China, in 2007, and a PhD degree in Pattern Recognition and Intelligent Systems from the University of Science and Technology of China, Hefei, China, in 2010. He has published more than 60 papers in international journals and conferences such as IEEE TPAMI, IEEE TNNLS, IEEE TCYB, IEEE TIP, IEEE TCSVT, IEEE TSMCS, KDD, and ACM MM. He is the Area Chair, Senior PC Member, or PC Member of many conferences such as NeurIPS and ICML. He is an Associate Editor of IEEE Transactions on Circuits and Systems for Video Technology (T-CSVT), Artificial Intelligence Review, Neural Networks, and Neurocomputing. His research interests include machine learning, pattern recognition, and image processing.
\end{IEEEbiography}

\vspace{-0.1cm}
\begin{IEEEbiography}[{\includegraphics[width=1in,height=1.25in,clip,keepaspectratio]{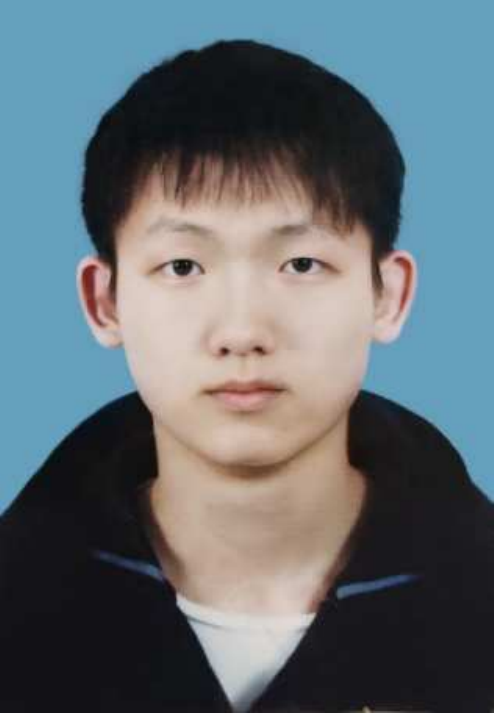}}]
{Chengze Jiang} (Student Member, IEEE) received the B.E. degree in software engineering from Guangdong Ocean University, Zhanjiang, China, in 2019. and M.Agr. degree in agricultural engineering and information technology from Guangdong Ocean University, Zhanjiang, China, in 2022. He is currently pursuing the Ph.D. degree in School of Cyber Science and Engineering, Southeast University, Nanjing, China. His current research interests include neural networks, optimization, and adversarial attack.
\end{IEEEbiography}

\vspace{-0.1cm}
\begin{IEEEbiography}[{\includegraphics[width=1in,height=1.25in,clip,keepaspectratio]{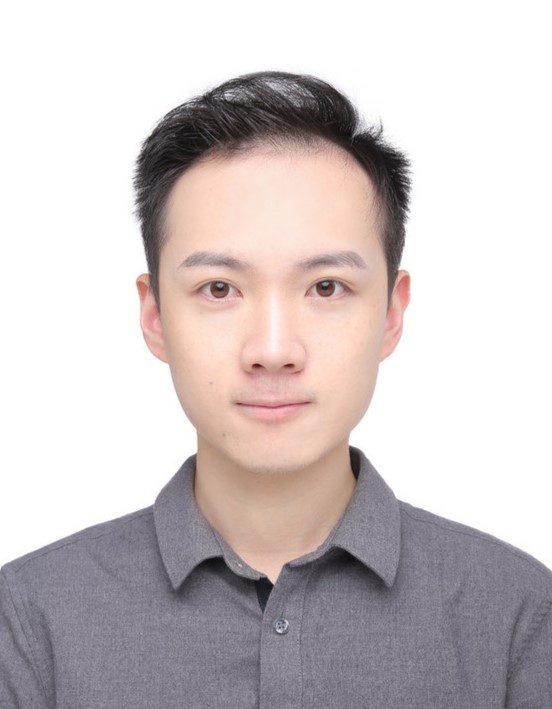}}]
{Minjing Dong} is currently an assistant professor at the department of computer science, City University of Hong Kong. He received a BS degree in Software Engineering from Dalian University of Technology, a BS degree in Information Technology from University of Sydney, a M.Phil. degree in Engineering and Information Technology from University of Sydney, and a Ph.D. degree in Engineering and Information Technology from University of Sydney. His research interests include adversarial robustness, model calibration, efficient neural network, human motion analytics, and generative models. He has published more than 25 papers in top-tier conferences and journals, such as NeurIPS, ICML, CVPR, AAAI, ICLR, TPAMI, TNNLS, TIP, and TMM. He received AAAI 2023 Distinguished Paper Award. He is the Area Chair and PC Member of many conferences and journals, such as NeurIPS, ICML, CVPR, ICCV, and ECCV.
\end{IEEEbiography}

\vspace{-0.1cm}
\begin{IEEEbiography}[{\includegraphics[width=1in,height=1.25in,clip,keepaspectratio]{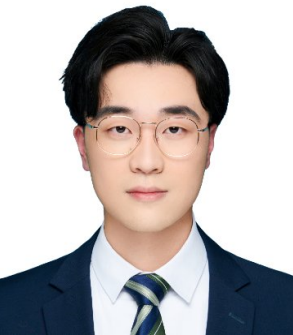}}]
{Kun Tong} received the B.E. degree in information security from Jiangsu University, Zhenjiang, China, in 2021. and M.S. degree in Cyber Science and Engineering from Southeast University, Nanjing, China, in 2024. His current research interests include neural networks and adversarial attack.
\end{IEEEbiography}

\vspace{-0.1cm}
\begin{IEEEbiography}[{\includegraphics[width=1in,height=1.25in,clip,keepaspectratio]{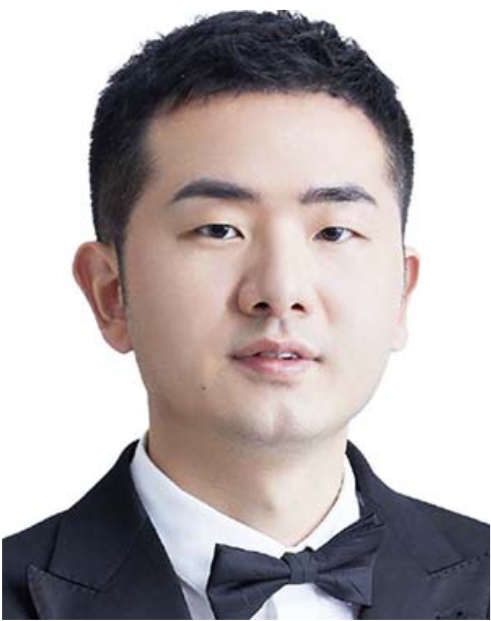}}]
{Xinli Shi} (Senior Member, IEEE) received the B.S. degree in software engineering, the M.S. degree in applied mathematics, and the Ph.D. degree in control science and engineering from Southeast University, Nanjing, China, in 2013, 2016, and 2019, respectively. In 2018, he held a China Scholarship Council Studentship for one-year study with the University of Royal Melbourne Institute of Technology, Melbourne, VIC, Australia. He is currently an Associate Professor with the School of Cyber Science and Engineering, Southeast University. His current research interests include distributed optimization, nonsmooth analysis, and network control systems. Dr. Shi was the recipient of the Outstanding Ph.D. Degree Thesis Award from Jiangsu Province, China.
\end{IEEEbiography}

\vspace{-0.1cm}
\begin{IEEEbiography}[{\includegraphics[width=1in,height=1.25in,clip,keepaspectratio]{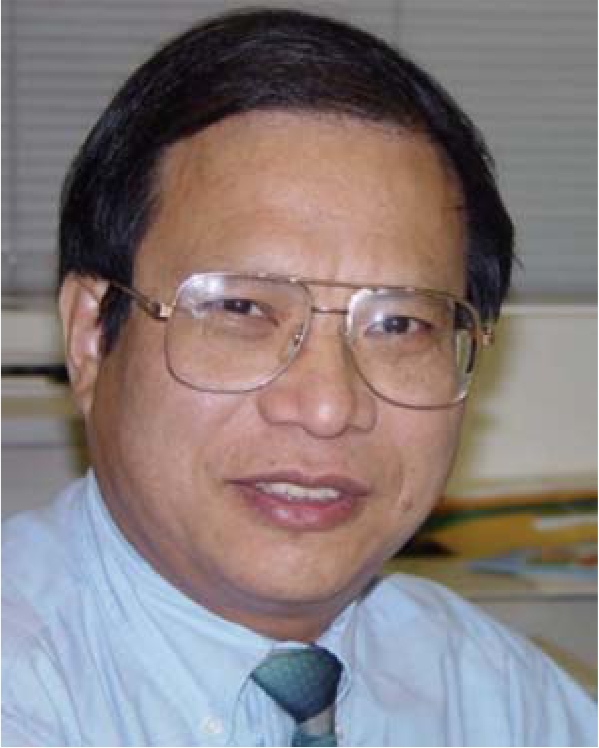}}]{Yuan Yan Tang} (Life Fellow, IEEE) received the BSc degree in electrical and computer engineering from Chongqing University, Chongqing, China, the MEng degree in electrical engineering from the Beijing Institute of Posts and Telecommunications, Beijing, China, and the PhD degree in computer science from Concordia University, Montreal, QC, Canada. He is currently an Emeritus chair professor with the Faculty of Science and Technology, University of Macau, and a professor/adjunct professor/honorary professor with several institutes, including Chongqing University, China, Concordia University, and Hong Kong Baptist University, Hong Kong. His current interests include wavelets, pattern recognition, and image processing. He is the founder and the chair of Pattern Recognition Committee, IEEE SMC. He has serviced as the general chair, program chair, and committee member of many international conferences. He is the founder and the general chair of the International Conferences on Wavelets Analysis and Pattern Recognition (ICWAPRs) series. He is the founder and editor-in-chief of the {\it International Journal of Wavelets}, {\it Multiresolution and Information Processing (IJWMIP)}. He is a fellow of IAPR.
\end{IEEEbiography}

\vspace{-0.1cm}
\begin{IEEEbiography}[{\includegraphics[width=1in,height=1.25in,clip,keepaspectratio]{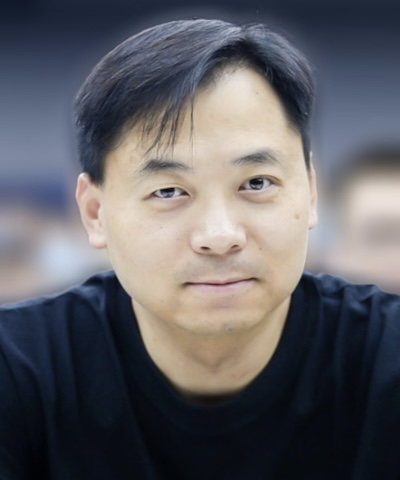}}]{Dacheng Tao} (Fellow, IEEE) is currently a Distinguished University Professor in the College of Computing \& Data Science at Nanyang Technological University. He mainly applies statistics and mathematics to artificial intelligence and data science, and his research is detailed in one monograph and over 200 publications in prestigious journals and proceedings at leading conferences, with best paper awards, best student paper awards, and test-of-time awards. His publications have been cited over 112K times and he has an h-index 160+ in Google Scholar. He received the 2015 and 2020 Australian Eureka Prize, the 2018 IEEE ICDM Research Contributions Award, and the 2021 IEEE Computer Society McCluskey Technical Achievement Award. He is a Fellow of the Australian Academy of Science, AAAS, ACM and IEEE.
\end{IEEEbiography}

\end{document}